\documentclass[journal]{IEEEtran}
\usepackage{amsmath,amsfonts}
\usepackage{algorithm2e}
\usepackage{array}
\usepackage{textcomp}
\usepackage{stfloats}
\usepackage{url}
\usepackage{verbatim}
\usepackage{graphicx}
\usepackage{adjustbox}
\usepackage{cite}
\usepackage{booktabs}
\usepackage{braket} 
\usepackage{soul}
\usepackage{placeins}
\usepackage[dvipsnames]{xcolor}
\usepackage{xurl}
\usepackage[hidelinks,breaklinks=true]{hyperref}
\usepackage{cleveref}
\usepackage{tabularx}
\usepackage[utf8]{inputenc}

\usepackage{colortbl} 

\SetKwComment{Comment}{/* }{ */}
\RestyleAlgo{ruled}

\usepackage{xcolor} 

\usepackage[switch]{lineno}
\usepackage{multicol}
\usepackage{multirow}
\usepackage{mwe}
\usepackage{float}
\usepackage{balance}
\usepackage{lipsum}
\usepackage{balance}

\ifCLASSINFOpdf
\else
\fi

\begin{document}

\author{
    \IEEEauthorblockN{Luigi Russo, \IEEEmembership{Student Member, IEEE}, 
    Anabella Ferral, \IEEEmembership{Senior Member, IEEE}, 
    Silvia Liberata Ullo, \IEEEmembership{Senior Member, IEEE}, 
    and Paolo Gamba, \IEEEmembership{Fellow, IEEE}}
    
    \thanks{Luigi Russo and Paolo Gamba are with the Department of Electrical, Computer and Biomedical Engineering, University of Pavia, 27100 Pavia, Italy (e-mail: luigi.russo02@universitadipavia.it; paolo.gamba@unipv.it).}

   \thanks{Anabella Ferral is with Instituto Gulich, UNC-CONAE-CONICET, Córdoba, Argentina (e-mail: aferral@unc.edu.ar).}

    \thanks{Silvia Liberata Ullo is with the Department of Engineering, University of Sannio, 82100 Benevento, Italy (e-mail: ullo@unisannio.it).}
    
}

\title{Multi-Sensor Mapping of Vulnerable Urban Settlements Using SAR, Multispectral, and Hyperspectral Imagery: A Case Study in Córdoba, Argentina}

\maketitle
 
\begin{abstract}
Informal settlements represent a major urban challenge in rapidly expanding cities, yet their identification from Earth Observation (EO) data remains difficult because of their heterogeneous appearance and incomplete official inventories. This work presents a multi-sensor deep learning (DL) framework for slum-likelihood mapping in Córdoba, Argentina, integrating high-resolution PlanetScope multispectral (MS) imagery, COSMO-SkyMed (CSK) Synthetic Aperture Radar (SAR) data, and medium-resolution PRISMA hyperspectral (HS) observations. The problem is formulated as a patch-level classification task using the official Registro Nacional de Barrios Populares (ReNaBaP) inventory as reference, and the models are evaluated through four geographically partitioned folds.
SAR-only and MS-only baselines, their configurations with PRISMA HS support, and early fusion (EF), middle fusion (MF), and late fusion (LF) strategies are systematically compared. Results show that LF+HS provides the best overall balance between classification performance and spatial selectivity, while PRISMA contributes complementary spectral information alongside the higher-resolution MS and SAR representations. Beyond the standard evaluation against ReNaBaP, an external municipal vulnerability layer is used to interpret detections outside the official polygons, showing that several apparent false positives overlap broader vulnerable urban areas. Thermal analysis further shows that ReNaBaP settlements exhibit significantly higher surface temperatures than their immediate surroundings during a heatwave event, indicating localised surface-heat amplification. Taken together, these results suggest that multi-sensor EO fusion can support both the mapping of ReNaBaP settlements and the interpretation of broader urban vulnerability patterns.
\end{abstract}


\IEEEpeerreviewmaketitle

\section{Introduction}\label{sec:intro}
Urban growth has changed not only the size of many cities, but also the way housing vulnerability appears at their margins and within consolidated urban fabrics. In contexts where planning systems and infrastructure provision do not keep pace with population growth, informal settlements, commonly referred to as ``slums'', often become part of the urban landscape. The United Nations defines a slum household as one lacking one or more basic conditions, including durable housing, sufficient living space, access to safe water and sanitation, and secure tenure \cite{UNHabitat2003}. More than one billion people are estimated to live in slum-like conditions worldwide \cite{UNHabitat2022}, and this number is expected to increase further in the coming decades \cite{Yang2026}. Although informality is a global phenomenon, sub-Saharan Africa illustrates particularly clearly how informal settlements can become a major component of rapid urban expansion, often emerging along fast-growing urban peripheries \cite{Buttner2025}.

The relevance of informal-settlement mapping goes beyond the delineation of built-up areas. These settlements are often associated with inadequate housing, limited infrastructure, insecure tenure, environmental exposure, overcrowding and reduced access to basic services. Their spatial identification is therefore important for monitoring Sustainable Development Goal (SDG) 11 and for supporting urban planning, risk reduction and resource allocation \cite{UN2015}. Censuses, field surveys and official settlement inventories remain indispensable because they provide socio-economic and institutional information that satellite data alone cannot capture. At the same time, these sources are expensive to update, may be produced at irregular intervals and can miss recent changes, especially in peripheral or rapidly transforming urban areas. Moreover, differences in definitions and survey protocols across countries and institutions can limit the comparability of settlement maps and official statistics.

Remote sensing (RS) can complement these sources by providing spatially explicit and repeatable observations of the built-up environment. Satellite imagery can describe several visible properties of informal settlements, including building density, roof materials, street layout, vegetation cover and local morphology. These properties are useful for separating informal settlements from the surrounding urban fabric, but they do not behave as fixed or universal indicators \cite{Kuffer2016,Mahabir2018}. Their relevance depends on the local urban context: informal settlements may differ substantially across cities, and even within the same city their form, materials and spatial organisation can vary from one neighbourhood to another \cite{Taubenbock2018,Friesen2018,GramHansen2019}. For this reason, image-based patterns cannot be interpreted independently of the socio-spatial context in which they occur.

Early RS approaches to informal-settlement mapping mainly relied on handcrafted features designed to describe the visible structure of these areas. Texture measures, morphological descriptors, spectral indices and object-based image analysis were used to capture patterns such as dense built-up fabric, irregular street layouts, small roof structures and limited vegetation. These aspects are widely recognised as key image-based indicators of informal settlements, with morphology, texture and spatial scale playing a central role in their distinction from the surrounding urban fabric \cite{Kuffer2016}. For instance, grey-level co-occurrence matrix descriptors and other spatial features extracted from very-high-resolution (VHR) imagery were used to represent local settlement structure \cite{Kuffer2016_2}. Although these methods represented an important step forward, their effectiveness was closely tied to the quality and suitability of the selected features. Since such descriptors are manually designed, they may be sensitive to illumination conditions, acquisition geometry, spatial resolution and local urban morphology. As a result, traditional machine-learning (ML) pipelines often performed well in the specific settings for which they were calibrated, but their transferability to different image conditions or settlement forms remained limited \cite{Tuia2016}.

Deep learning (DL) changed this perspective by replacing handcrafted descriptors with learned representations. Convolutional neural networks (CNNs) and fully convolutional networks (FCNs) have been successfully applied to VHR imagery and have improved the detection of informal settlements compared with traditional ML methods \cite{Mboga2017,Helber2019FDL}. Transfer learning and semantic segmentation models further enabled more detailed delineation of slum areas, even when labelled data were limited \cite{Wurm2019}. Other studies investigated multi-scale architectures and change-detection strategies to better capture the irregular geometry and temporal evolution of informal settlements \cite{Liu2019}. Related work has also shown that CNNs can identify fine-scale deprivation patterns beyond strictly defined slum boundaries, supporting a broader interpretation of satellite-derived vulnerability indicators \cite{Wang2019Deprivation,ArribasBel2017}. Nevertheless, DL models remain affected by the same fundamental issue: they may learn context-specific associations between visual patterns, sensor properties and annotation practices. This has motivated research on transfer learning, semi-supervised learning and domain adaptation in RS \cite{ElMoudden2024}, as well as recent approaches aimed at improving generalisation under limited supervision and across heterogeneous geographic contexts \cite{Lee2025}.

In parallel, the increasing availability of open-access Earth Observation (EO) data and large-scale computational resources has made large-area mapping increasingly feasible. Recent studies have generated slum-probability or deprivation maps over hundreds of cities using medium-resolution satellite imagery, moving informal-settlement mapping towards broader multi-city frameworks. These works have also introduced uncertainty-aware strategies, such as ensemble modelling and test-time dropout, to express confidence across heterogeneous urban contexts \cite{Stark2025}. Similarly, scalable approaches based on freely available Sentinel-2 imagery have mapped deprived urban areas across multiple African cities, showing the practical value of models that can be applied when detailed local inventories are missing \cite{Owusu2024}. Nevertheless, transferability remains a critical issue in informal-settlement mapping, since settlement morphology, construction materials, infrastructure deficits and official labelling practices may change substantially from one urban context to another \cite{Owusu2021}.

These considerations motivate the need for local analyses. Large-scale studies are essential for regional or global monitoring, but they often involve compromises in spatial detail, reference-data quality and contextual interpretation. City-specific studies can therefore provide complementary insight by examining how model predictions relate to local inventories, municipal vulnerability layers and physically interpretable urban conditions.

A second line of research concerns the use of multiple sensors. Within the optical domain, multispectral (MS) imagery has been widely used for informal-settlement mapping because it provides spectral and textural information related to vegetation cover, exposed soil, roof materials and local urban morphology \cite{Kuffer2016,Mahabir2018}. Hyperspectral (HS) imagery extends this information by sampling the reflected signal with a much finer spectral resolution, allowing material-related variability to be described in greater detail \cite{Kahraman2021HSIFusion,Matarira2023}. In the case of PRISMA, this spectral richness is provided at medium spatial resolution for the HS cube, whereas the higher-resolution panchromatic band does not contain hyperspectral information and is therefore not used in this work. SAR data provide a complementary, non-optical observation source, since they are sensitive to surface roughness, geometry and built-up structure, and have been shown to provide useful spatial information for slum mapping \cite{Wurm2017SAR}. More broadly, multi-sensor fusion has become central in RS because different sensors capture complementary physical properties of the same urban scene.
Recent studies have combined satellite imagery and street-view data to better interpret informal urban environments \cite{Niu2025}. Other works have shown that combining MS imagery with ancillary environmental and spatial descriptors can improve DL-based slum mapping \cite{GASlumNet2024}, while projects such as IDEAtlas demonstrate how Sentinel-1, Sentinel-2 and ancillary geospatial layers can be integrated with stakeholder validation to support deprived-urban-area mapping \cite{Tareke2025}.

Despite these advances, systematic studies jointly exploiting high-resolution MS imagery, SAR data and HS observations for informal-settlement mapping remain limited. In particular, three aspects are still insufficiently explored. First, the specific contribution of each sensing modality and the effect of different fusion strategies are often difficult to disentangle. Second, model evaluation is commonly performed against official settlement inventories that may be incomplete, outdated or dependent on local administrative definitions. Under this type of evaluation, detections outside official polygons are usually counted as false positives, although some of them may correspond to vulnerable urban areas that are not represented in the adopted inventory. A further limitation is that many DL-based approaches provide limited physical interpretability, making it difficult to relate the predicted patterns to urban characteristics such as construction materials, settlement density, vegetation scarcity or environmental exposure.

These limitations are particularly relevant because informal settlements are not defined solely by image appearance. Their morphology, materials, infrastructure conditions and institutional recognition may vary substantially across cities and countries. Consequently, neighbourhoods with similar environmental configurations may be labelled differently depending on the adopted inventory, while other vulnerable areas may remain outside official settlement registers. In this context, local multi-sensor analyses are useful not only to improve classification performance, but also to examine how EO-based outputs relate to the urban, institutional and environmental conditions of the specific study area.

This study addresses these issues through a multi-sensor framework designed to map informal and vulnerable settlements in Córdoba, Argentina. In Argentina, the main official reference for this purpose is the Registro Nacional de Barrios Populares (ReNaBaP), which identifies, characterises and georeferences \emph{barrios populares} across the country. The inventory supports socio-urban integration policies and reports 6,467 officially registered \emph{barrios populares} in its latest available version \cite{RENABAP2024}. These areas are generally understood as grouped or contiguous settlements where residents face limited access to basic services, such as water supply or electricity.

The objective of this work is to investigate how complementary EO observations and external contextual information can support the mapping and interpretation of vulnerable urban settlements within a specific urban system. Córdoba represents a relevant test case because ReNaBaP provides an official georeferenced inventory of \emph{barrios populares}, while the Municipality of Córdoba vulnerability layer (CBA) provides an external description of broader neighbourhood-scale vulnerability conditions. This makes it possible to assess not only whether the model identifies ReNaBaP settlements, but also whether detections outside ReNaBaP are spatially consistent with wider vulnerable urban contexts.

The proposed framework combines PlanetScope (PS) MS imagery, COSMO-SkyMed (CSK) SAR data and PRISMA HS observations within a DL workflow based on geographically partitioned training and evaluation areas. MS and SAR data are used to capture complementary spectral, textural and structural information, while PRISMA observations are incorporated as compact material-sensitive spectral descriptors. The experiments compare SAR-only and MS-only baselines, their corresponding single-spatial-backbone configurations with PRISMA HS support, and EF, MF, and LF strategies for SAR--MS integration, with the fusion strategies evaluated both with and without HS support. This comparison allows the contribution of each modality to patch-level classification and city-wide slum-likelihood mapping to be assessed.

Beyond the standard evaluation against the official ReNaBaP inventory, we also perform a contextual assessment using the external CBA vulnerability layer. This analysis examines whether detections outside the ReNaBaP polygons spatially overlap broader vulnerable urban areas identified by the Municipality of Córdoba, without treating the CBA layer as an additional reference inventory or reclassifying the corresponding false positives. A separate Landsat-8 thermal analysis is also performed to investigate surface-temperature differences between ReNaBaP settlements and their immediate surroundings during a heatwave event.

In this context, the main contribution of this work is a local contextual assessment of multi-sensor EO data for slum-likelihood mapping in Córdoba. Specifically, this study:
\begin{itemize}
\item implements and evaluates a multi-sensor DL framework combining high-spatial-resolution PlanetScope MS imagery, COSMO-SkyMed SAR backscatter and PCA-reduced PRISMA HS descriptors for patch-level slum-likelihood classification, using ReNaBaP as the reference inventory;
\item compares SAR-only and MS-only baselines, their corresponding single-spatial-backbone configurations with HS support, and EF, MF, and LF strategies for SAR--MS integration under a spatial OOF evaluation protocol;
\item performs a contextual assessment of detections outside ReNaBaP using the external CBA vulnerability layer, without treating it as an additional informal-settlement reference inventory;
\item complements the mapping results with an additional thermal analysis based on Landsat-8 TIRS data, providing a complementary interpretation of the surface-temperature patterns associated with ReNaBaP settlements.
\end{itemize}

\noindent The remainder of this paper is organised as follows. Section~\ref{sec:study_area_data} describes the Córdoba study area, the EO datasets, the reference and contextual data, and the construction of the spatially aligned multi-sensor dataset. Section~\ref{sec:methodology} presents the proposed learning framework, including the PRISMA HS embedding strategy, the SAR- and MS-based configurations, the multi-sensor fusion schemes, the training protocol and the OOF city-wide mapping procedure. Section~\ref{sec:results_discussion} reports the experimental results, including patch-level classification performance, city-wide slum-likelihood mapping, contextual interpretation with the CBA vulnerability layer, vulnerability-discrimination analysis, and thermal interpretation during a heatwave event. Finally, Section~\ref{sec:conclusions} concludes the paper and discusses possible future developments.

\begin{figure*}[t]
\centering
\includegraphics[width=0.9\textwidth]{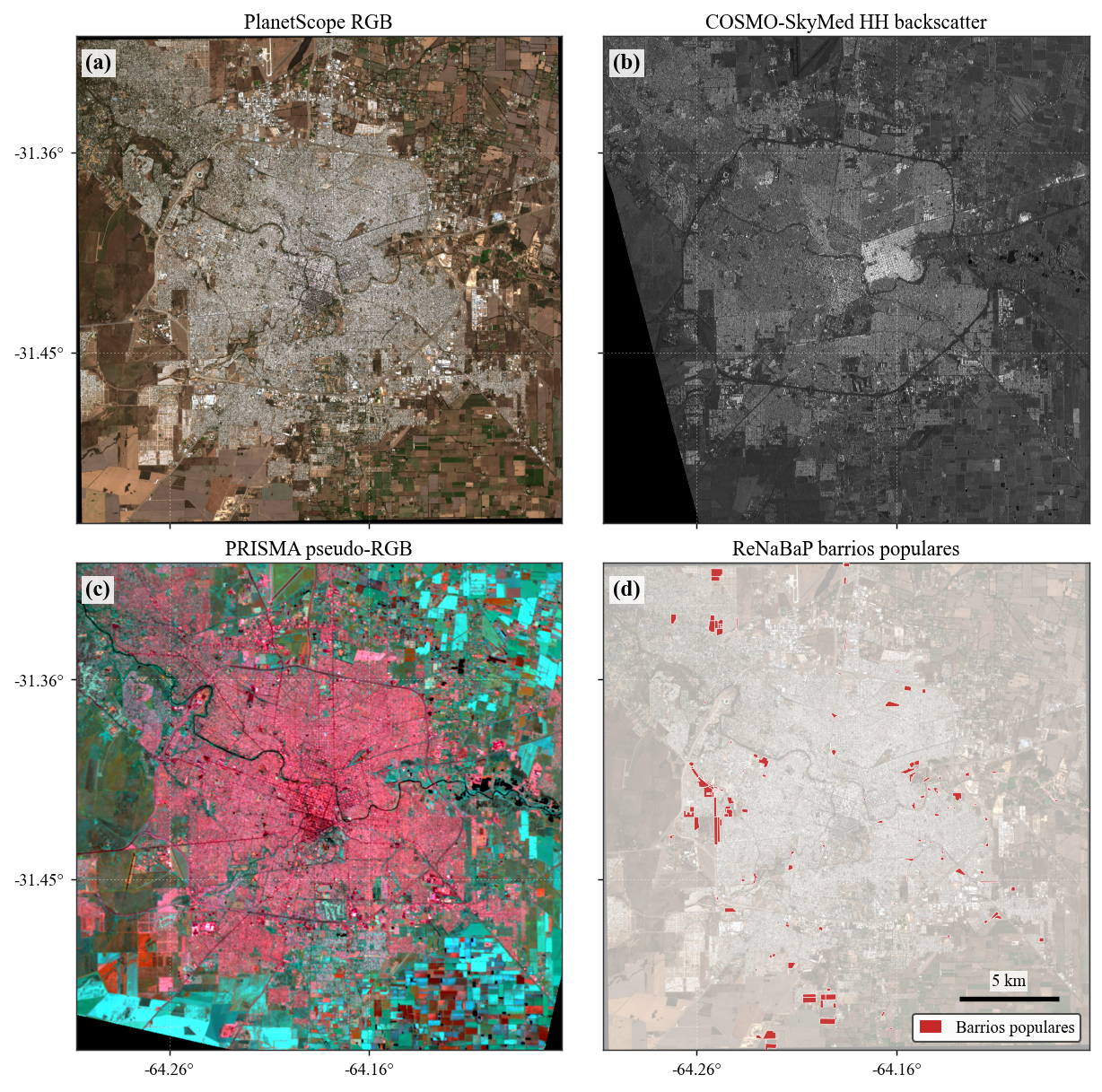}
\caption{
Study area in Córdoba, Argentina, and EO data used in this work.
The figure shows (a) PlanetScope RGB imagery acquired on 19 July 2025, (b) COSMO-SkyMed HH SAR backscatter acquired on 28 July 2025, (c) PRISMA false-colour composite generated from selected hyperspectral bands and acquired on 28 March 2021, and (d) the ReNaBaP polygons overlaid on PlanetScope RGB imagery.
All panels refer to the same spatial extent and provide the basis for model training, prediction, and contextual evaluation.
}
\label{fig:study_area}
\end{figure*}

\section{Study Area, Data and Dataset Construction}
\label{sec:study_area_data}

\subsection{Study Area and EO Data}
\label{subsec:study_area_eo_data}

The study area is the city of Córdoba, Argentina, located in the central part of the country and representing one of its main urban, economic and academic centres. Córdoba is the second-largest city in Argentina, with an urban jurisdiction of about 576 km$^2$ and a population of approximately 1.5--1.6 million inhabitants \cite{MobiliseYourCityCordoba2025,MetropolisCordoba}. Its urban fabric includes a consolidated central core, industrial and residential districts, major transport corridors and expanding peri-urban neighbourhoods. Previous urban-expansion analyses indicate that the built-up extent of Córdoba increased substantially between 2001 and 2014, with growth occurring along multiple peripheral directions \cite{AtlasUrbanExpansionCordoba}. This spatial heterogeneity makes the city a relevant case study for analysing how slum-like and vulnerable settlements appear under different morphological and infrastructural conditions.

The EO dataset used in this study combines high-resolution MS imagery, SAR data and HS observations over the Córdoba urban area. These sources were selected because they describe complementary physical properties of the urban environment. PS imagery was provided by Planet Labs PBC \cite{PlanetScopeDocs}. The PS product consists of eight surface-reflectance bands and covers the study area at approximately 3 m spatial resolution. CSK data consist of a terrain-geocoded SAR product acquired in polarisation HH mode, contributing information on surface roughness, built-up geometry and structural properties of the urban fabric. PRISMA HS observations provide material-sensitive spectral information across the VNIR--SWIR range through an original cube composed of 230 spectral bands at 30 m spatial resolution.

The EO, reference and contextual datasets used in this study are summarised in Table~\ref{tab:data_sources}. PS imagery was used as the reference grid for high-resolution patch extraction. CSK data were provided by the Italian Space Agency (ASI) and consist of a terrain-geocoded Level-1D product acquired in HH polarisation. Before patch extraction, the CSK and PS rasters were reprojected to a common coordinate reference system (CRS) and resampled onto aligned 3~m grids. PRISMA HS data, also provided by ASI, were processed at their native spatial resolution of 30~m and subsequently transformed into compact spectral descriptors.

In addition to the EO datasets used for model training and inference, the Landsat~8 OLI/TIRS Collection~2 Level-2 surface-temperature product acquired on 3 January 2022 was used exclusively for thermal interpretation. Specifically, the surface-temperature band ($\mathrm{ST_{B10}}$) was converted from the distributed scaled digital numbers to degrees Celsius according to:

\begin{equation}
T_{\mathrm{C}} = \mathrm{ST_{B10}} \times 0.00341802 - 124.15.
\end{equation}

The acquisition coincided with a severe heatwave affecting central Argentina and was therefore selected to investigate surface-temperature differences between ReNaBaP settlements and their immediate surroundings \cite{SMN2025}. The thermal data were not included in model training, threshold selection or city-wide inference.

Fig.~\ref{fig:study_area} provides an overview of the common analysis extent and of the EO inputs used in this work.

\subsection{Reference and Contextual Data}
\label{subsec:reference_data}

The reference settlement information is provided by the Registro Nacional de Barrios Populares (ReNaBaP), which maps officially recognised \emph{barrios populares} in Argentina  \cite{RENABAPMap}, hereafter referred to as ReNaBaP settlements. For the Córdoba study area, the dataset is provided as a georeferenced polygon layer describing the spatial extent of officially recognised settlements. The layer includes administrative and descriptive attributes associated with each settlement polygon, such as settlement identifiers, institutional registration information and metadata related to the official inventory process. In this study, ReNaBaP polygons are used as the institutional baseline for defining the supervised mapping task and for deriving the binary patch-level labels adopted during training and validation.

However, ReNaBaP is not treated as an exhaustive representation of urban vulnerability. Official inventories may be incomplete, updated at irregular intervals or limited to settlements that have already been formally recognised by public institutions. Consequently, vulnerable neighbourhoods may exist outside the official inventory while still sharing morphological, infrastructural or socio-spatial characteristics with registered \emph{barrios populares}.

To support a broader contextual interpretation of the resulting slum-likelihood maps, we also use the Municipality of Córdoba Vulnerability Layer (CBA). The dataset is organised at neighbourhood scale and assigns each mapped unit to one of two vulnerability classes: high vulnerability or low vulnerability. In the study area, it includes 186 neighbourhood units, of which 122 are classified as high vulnerability and 64 as low vulnerability. Whereas ReNaBaP delineates officially registered \emph{barrios populares}, the CBA layer describes broader neighbourhood-scale vulnerability conditions.

The CBA layer is not used during model training or label generation, but only for contextual interpretation of the model outputs. Specifically, it allows us to examine whether areas with high slum-likelihood scores detected outside the official ReNaBaP inventory are spatially associated with broader vulnerable neighbourhoods identified by the municipality, rather than representing only classification errors. In this way, the CBA layer provides an external source of contextual information for interpreting the resulting multi-sensor slum-likelihood maps.

Fig.~\ref{fig:study_area}(d) shows the ReNaBaP polygons overlaid on the PlanetScope RGB imagery covering the study area.

\begin{table*}[t]
\centering
\caption{EO, reference and contextual datasets used in this study.}
\label{tab:data_sources}
\scriptsize
\setlength{\tabcolsep}{3.5pt}
\renewcommand{\arraystretch}{1.18}
\resizebox{\textwidth}{!}{%
\begin{tabular}{llllll}
\hline
Data source & Category & Acquisition date / version & Product / attributes & Spatial resolution / unit & Role in the workflow \\
\hline
COSMO-SkyMed & SAR image & 28-07-2025 & L1D GTC, single-pol HH backscatter & $\sim$3 m & High-resolution SAR input \\
PlanetScope & Multispectral image & 19-07-2025 & Surface reflectance, 8 bands & $\sim$3 m & High-resolution MS input \\
PRISMA & Hyperspectral image & 28-03-2021 & L2D surface reflectance, original 230-band cube & 30 m & PCA-reduced spectral descriptors \\
Landsat-8 TIRS & Thermal image & 03-01-2022 & Collection 2 Level-2 Band 10 surface temperature product & 100 m & Surface-temperature comparison \\
ReNaBaP & Reference polygons & 2025 & Official \emph{barrios populares} inventory & Settlement polygons & Label derivation and reference-map comparison \\
CBA layer & Contextual polygons & 2025 & High-/low-vulnerability classes & Neighbourhood-scale units & Contextual interpretation \\
\hline
\end{tabular}%
}
\end{table*}

\subsection{Multi-Sensor Dataset Construction}
\label{subsec:dataset_construction}
\begin{figure*}[t]
    \centering
    \includegraphics[width=0.95\textwidth]{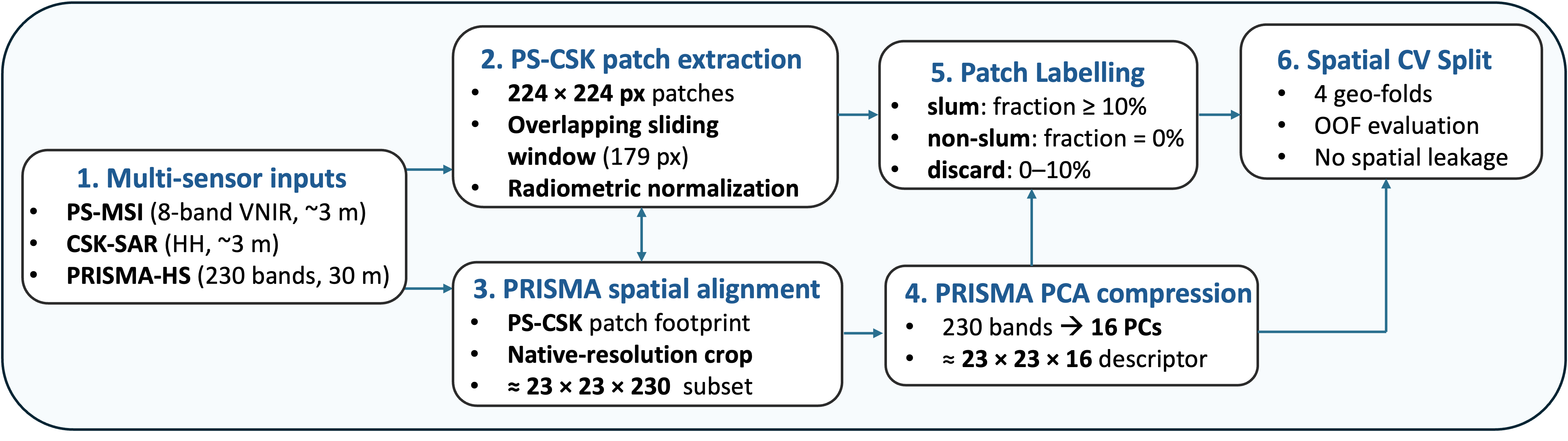}
    \caption{
    Multi-sensor dataset construction workflow. Spatially aligned PS and CSK patches are extracted from the high-resolution data, while corresponding PRISMA subsets are cropped at native resolution and compressed into 16 PCA components. ReNaBaP polygons are then used for binary patch labelling, and the resulting samples are split into four geographically partitioned folds for OOF training and evaluation.
    }
    \label{fig:dataset_construction}
\end{figure*}

The dataset construction procedure converts the EO inputs and ReNaBaP reference layer into a spatially aligned multi-sensor dataset, while the CBA layer is retained exclusively for subsequent contextual analysis. The overall preprocessing workflow is summarised in Fig.~\ref{fig:dataset_construction}.
Starting from spatially aligned PS and CSK data and geographically corresponding PRISMA observations, the procedure generates aligned MS--SAR patches, extracts corresponding native-resolution HS subsets, compresses PRISMA information into PCA-based descriptors, generates binary labels from the ReNaBaP inventory and defines geographically partitioned folds for OOF evaluation.

\subsubsection{PS--CSK Patch Extraction}
\label{subsubsec:ps_csk_patches}

PS was used as the reference grid for high-resolution patch extraction. Before patch generation, the CSK and PS rasters were reprojected to a common coordinate reference system and resampled onto aligned 3~m grids. The two datasets were then independently normalised to account for the different radiometric characteristics and value distributions of multispectral reflectance and SAR backscatter. Invalid or no-data pixels were excluded from the normalisation procedure.

The Córdoba study area was scanned using a sliding-window procedure with patches of $224 \times 224$ pixels and a stride of 179 pixels. Because the CSK raster had already been aligned with the PS reference grid, corresponding MS and SAR patches were extracted using the same pixel windows and therefore represented the same geographic footprints. Each retained sample contained an eight-channel PS patch and the corresponding single-channel CSK patch.

Patches affected by invalid or missing data in either modality were discarded. This ensured that each retained sample contained spatially aligned and valid MS and SAR observations over the same urban area.

The patch size was selected to capture local settlement characteristics, including built-up density, textural irregularity, vegetation context and neighbourhood morphology. The 179-pixel stride introduced controlled overlap between neighbouring patches, reducing border effects and supporting the subsequent reconstruction of spatially continuous city-wide slum-likelihood maps. The geographic footprint of each retained patch was preserved for label generation from the ReNaBaP mask and for projecting the model predictions back onto the PS reference grid.

\subsubsection{PRISMA Spatial Alignment and PCA Compression}
\label{subsubsec:prisma_processing}
\begin{figure}[t]
    \centering
    \includegraphics[width=\columnwidth]{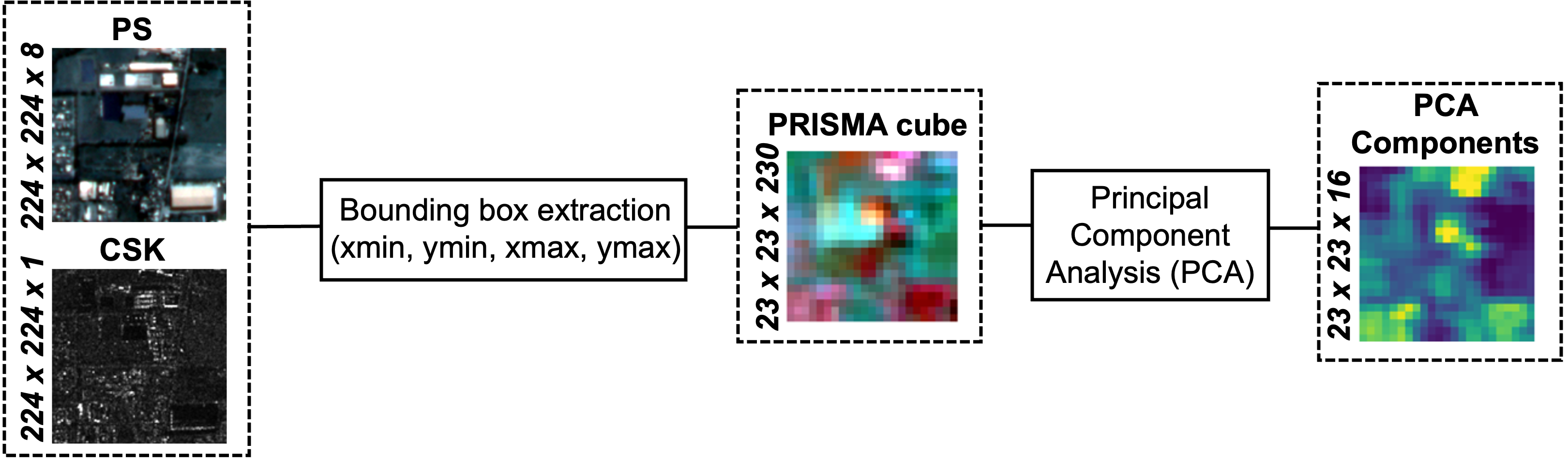}
    \caption{
    PRISMA alignment and PCA-based compression. Each PS--CSK patch footprint is used to crop a native-resolution PRISMA subset, which is reduced from 230 spectral bands to 16 principal components.
    }
    \label{fig:prisma_alignment}
\end{figure}

PRISMA HS data are processed differently from PS and CSK because of their coarser native spatial resolution. Rather than resampling the full HS cube to the high-resolution patch grid, PRISMA is exploited through spatially corresponding HS subsets. This design choice is motivated by the fact that the 30 m PRISMA resolution is too coarse to reliably capture fine-scale settlement morphology, roof patterns or detailed neighbourhood layout. Therefore, PRISMA is not used as an additional high-resolution spatial input, but mainly as a source of material-sensitive spectral information related to surface composition, exposed soil, vegetation conditions and moisture-related variability. For each PS patch, the same geographic bounding box used for the SAR and label extraction is used to crop the PRISMA cube at its native spatial resolution, as illustrated in Fig.~\ref{fig:prisma_alignment}.

For each retained PS--CSK patch, the corresponding PRISMA subset is cropped at native resolution using the same geographic footprint. The resulting HS subset has approximate size $230 \times 23 \times 23$, preserving the VIS--SWIR spectral information associated with the same urban area covered by the high-resolution inputs.

For each OOF split, a single PCA transformation was fitted using the spectral samples from the training folds and subsequently applied to the corresponding training and held-out PRISMA subsets. Let
\begin{equation}
    \mathbf{X}^{\mathrm{HS}}_i \in \mathbb{R}^{230 \times H \times W}
\end{equation}
denote the original PRISMA cube associated with patch $i$. The cube is reshaped into a spectral matrix
\begin{equation}
    \mathbf{X}^{\mathrm{HS}}_i
    \rightarrow
    \mathbf{M}_i \in \mathbb{R}^{(HW)\times230},
\end{equation}
projected onto the first $K=16$ principal components, and reshaped back into a compact HS descriptor:
\begin{equation}
    \tilde{\mathbf{X}}^{\mathrm{HS}}_i
    \in
    \mathbb{R}^{16 \times H \times W}.
\end{equation}

After projection, the PCA-reduced subsets retain their native spatial support and are subsequently spatially averaged and processed by the HS encoder during model training and inference. This dimensionality-reduction step reduces the computational cost while preserving the spectral information associated with each patch footprint.

\subsubsection{Patch Labelling Strategy}
\label{subsubsec:patch_labelling}

Patch labels are derived from a rasterised ReNaBaP mask generated on the reference raster grid. ReNaBaP polygons are first reprojected to the common CRS, checked for topological validity, and clipped to the analysis extent. The polygons are then rasterised into a binary mask, where pixels inside ReNaBaP settlements are assigned value 1 and all remaining pixels value 0.

For each extracted patch $i$, the corresponding label window is obtained using the same geographic footprint as the PS patch. The ReNaBaP-covered fraction is computed as the fraction of positive pixels in the label window:
\begin{equation}
    r_i =
    \frac{1}{N_i}
    \sum_{p=1}^{N_i}
    m_{i,p},
\end{equation}
where $m_{i,p} \in \{0,1\}$ denotes the rasterised ReNaBaP mask value for pixel $p$ in patch $i$, and $N_i$ is the number of pixels in the corresponding label window.

Following the 10\% threshold adopted by Stark et al. \cite{Stark2025}, patches with a ReNaBaP fraction greater than or equal to 10\% are labelled as ReNaBaP-positive patches, whereas patches with no ReNaBaP overlap are labelled as ReNaBaP-negative patches. Patches with a ReNaBaP fraction between 0\% and 10\% are discarded because they represent ambiguous boundary cases: 
\begin{equation}
    y_i =
    \begin{cases}
    1, & r_i \geq 0.10, \\
    0, & r_i = 0, \\
    \mathrm{discarded}, & 0 < r_i < 0.10.
    \end{cases}
\end{equation}

\begin{figure}[t]
    \centering
    \includegraphics[width=\columnwidth]{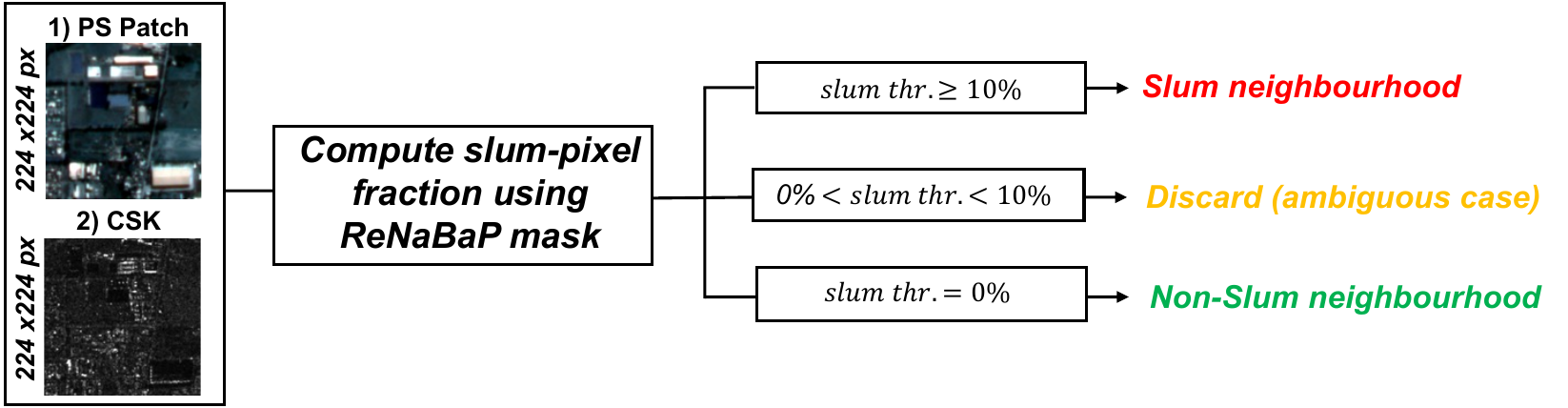}
    \caption{
    Patch labelling strategy based on the ReNaBaP-covered fraction within each patch. Patches with coverage $\geq$10\% are labelled as ReNaBaP-positive patches, patches with 0\% overlap are labelled as ReNaBaP-negative patches, and intermediate cases are discarded as ambiguous samples.
    }
    \label{fig:patch_labelling}
\end{figure}

This strategy limits boundary-related label noise by retaining only patches with clear ReNaBaP assignment. This labelling procedure is illustrated in Fig.~\ref{fig:patch_labelling}.

\subsubsection{Spatial Cross-Validation Split}
\label{subsubsec:spatial_cv}

To reduce geographical bias and limit spatial leakage, the study area is divided into four geographically partitioned folds, corresponding to different sectors of the Córdoba analysis extent. Each fold is used once as the held-out evaluation region, while the remaining folds are used for training. This 4-fold spatial CV design evaluates model performance over geographically partitioned areas rather than through random patch-level splitting.

\begin{figure}[t]
    \centering
    \includegraphics[width=\columnwidth]{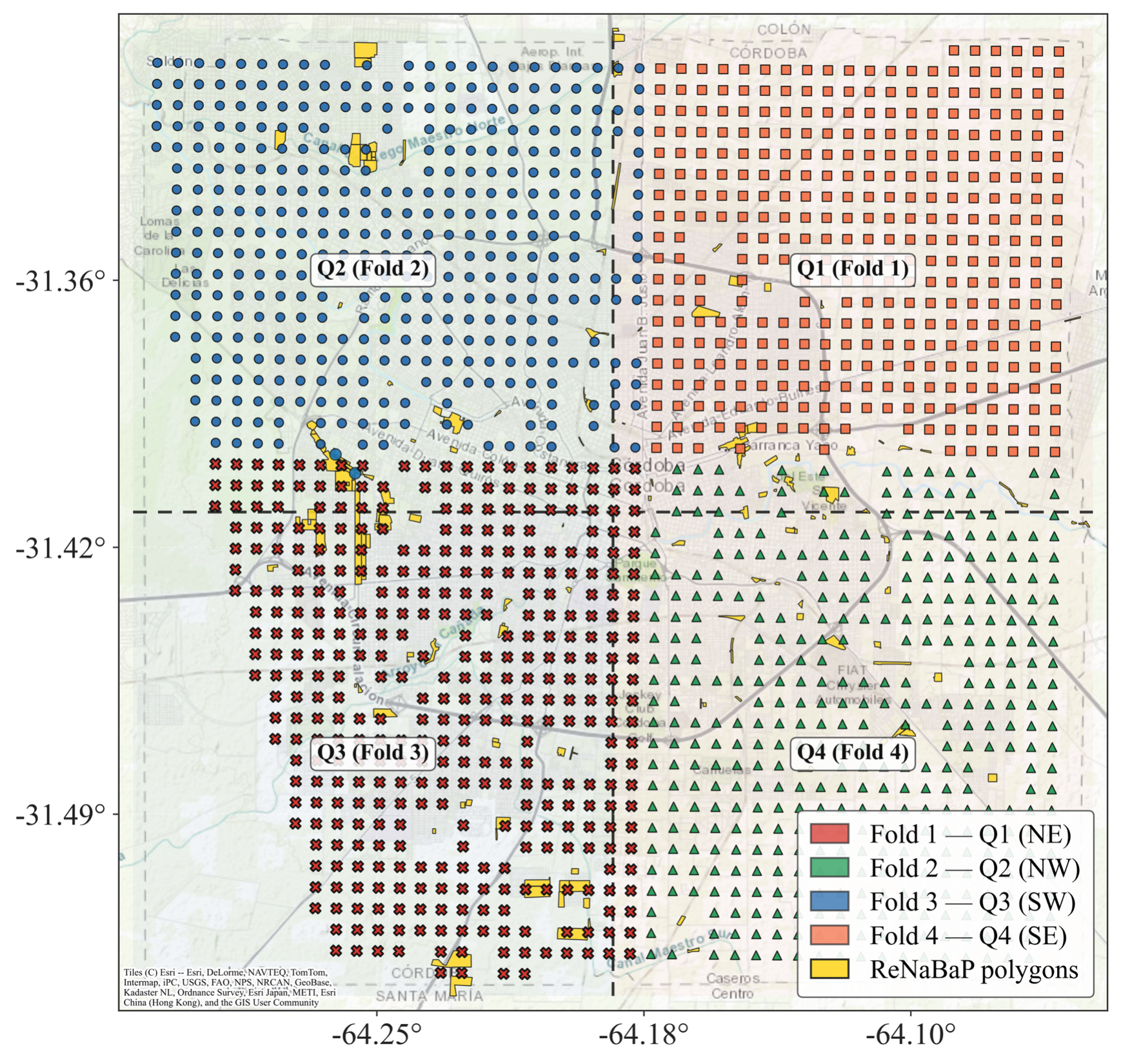}
    \caption{
    Spatial CV folds used for the Córdoba study area. The four sectors define geographically partitioned folds used for OOF evaluation, while ReNaBaP polygons indicate the spatial distribution of reference settlements across the city.
    }
    \label{fig:cv_folds}
\end{figure}

The resulting fold distribution is approximately balanced across the study area, as reported in Table~\ref{tab:spatial_cv_folds}. Fold~1 contains 383 patches, Fold~2 contains 403 patches, Fold~3 contains 385 patches and Fold~4 contains 363 patches, corresponding to 24.97\%, 26.27\%, 25.10\% and 23.66\% of the dataset, respectively, for a total of 1534 samples. The spatial partitioning is shown in Fig.~\ref{fig:cv_folds}.

The same fold structure is used for all modalities and fusion strategies. At each CV iteration, three folds are used for model training and the remaining fold is held out for evaluation. This OOF strategy generates predictions for geographically partitioned evaluation areas not used during the corresponding model-training iteration.

The resulting dataset comprises spatially aligned PS, CSK, and PRISMA-derived inputs, together with binary ReNaBaP-based labels and geographically separated evaluation areas. As described in Section~\ref{sec:methodology}, this dataset is used to train SAR- and MS-based models and SAR--MS fusion models, with and without PRISMA HS support, and to generate OOF city-wide slum-likelihood maps.

\section{Methodology}
\label{sec:methodology}

The proposed methodology formulates informal-settlement mapping as a binary local-scene classification problem, where each sample corresponds to a georeferenced image patch. Each retained sample is associated with a label $y_i \in \{0,1\}$, where $y_i=1$ denotes a ReNaBaP-positive patch and $y_i=0$ denotes a ReNaBaP-negative patch. The model outputs a scalar logit, which is converted into a slum-likelihood score and used for both patch-level evaluation and city-wide slum-likelihood mapping.

Each sample includes spatially aligned PS, CSK and PRISMA-derived inputs constructed as described in Section~\ref{sec:study_area_data}. Let $\mathbf{x}^{\mathrm{PS}}_i \in \mathbb{R}^{8 \times 224 \times 224}$ denote the PS multispectral patch, $\mathbf{x}^{\mathrm{SAR}}_i \in \mathbb{R}^{1 \times 224 \times 224}$ the corresponding CSK SAR patch, and $\tilde{\mathbf{x}}^{\mathrm{HS}}_i \in \mathbb{R}^{16 \times h \times w}$ the PCA-reduced PRISMA HS descriptor associated with the same geographic footprint.

\begin{figure*}[t]
\centering
\includegraphics[width=0.95\textwidth]{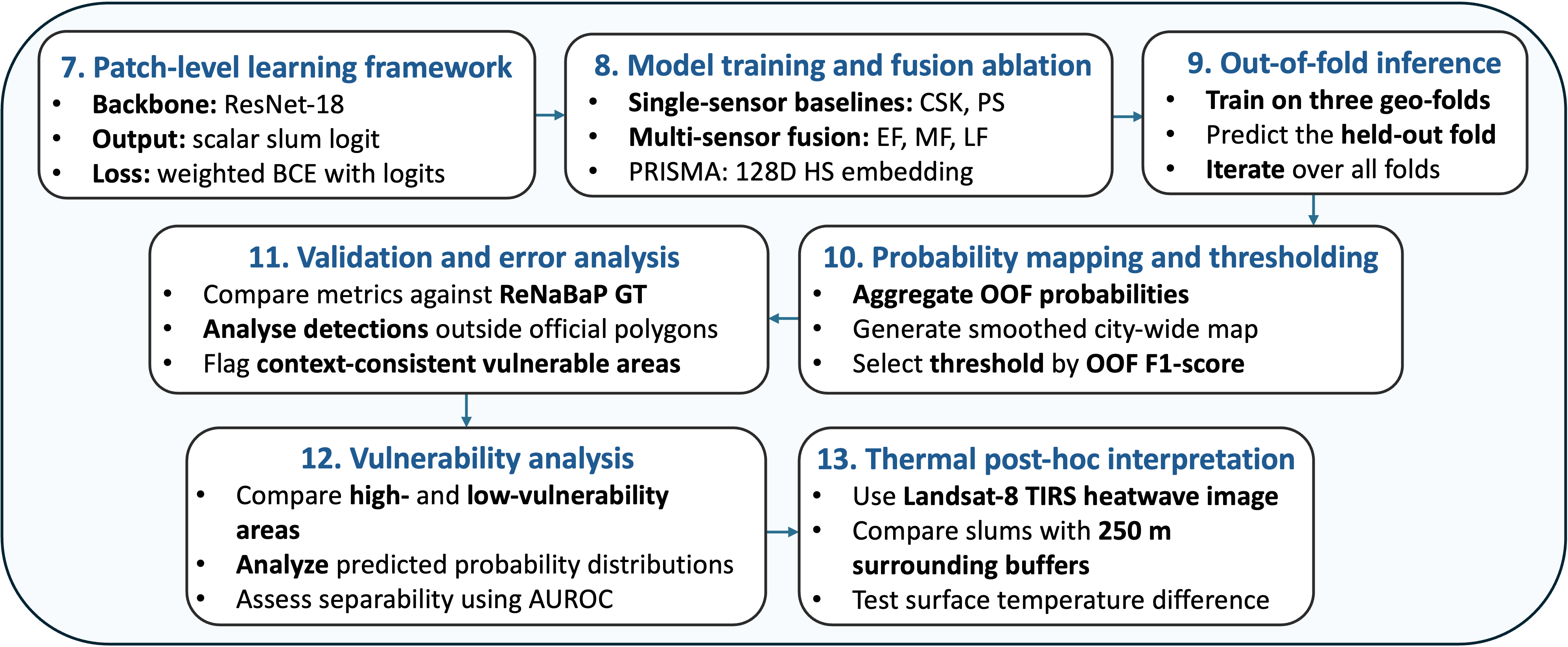}
\caption{
Methodological and interpretation workflow following the dataset construction stage.
The upper part summarises the patch-level learning framework, model-training ablation, spatial OOF inference and city-wide slum-likelihood mapping.
The lower part summarises the subsequent validation and interpretation steps, including ReNaBaP-based metric computation, contextual analysis of detections outside official polygons, vulnerability-discrimination assessment and complementary thermal interpretation.
{\small\emph{Note: The step numbering starts from 7 because this workflow continues the dataset-construction sequence shown in Fig.~\ref{fig:dataset_construction}.}}
}
\label{fig:methodology_workflow}
\end{figure*}

The overall workflow is summarised in Fig.~\ref{fig:methodology_workflow}. This section describes the learning framework, including the SAR- and MS-based configurations, the PRISMA embedding strategy, the multi-sensor fusion schemes, the training protocol, and the spatial OOF inference procedure. The resulting predictions are analysed in Section~\ref{sec:results_discussion} in terms of patch-level performance, city-wide slum-likelihood mapping, contextual assessment using the CBA vulnerability layer, and thermal interpretation.

\begin{table}[t]
\centering
\caption{Spatial CV fold distribution.}
\label{tab:spatial_cv_folds}
\begin{tabular}{ccccc}
\hline
\textbf{Fold 1} & \textbf{Fold 2} & \textbf{Fold 3} & \textbf{Fold 4} & \textbf{Total} \\
\hline
383 & 403 & 385 & 363 & 1534 \\
24.97\% & 26.27\% & 25.10\% & 23.66\% & 100.00\% \\
\hline
\end{tabular}
\end{table}

\begin{figure*}[t]
    \centering
    \begin{tabular}{c}
        \includegraphics[width=0.78\textwidth]{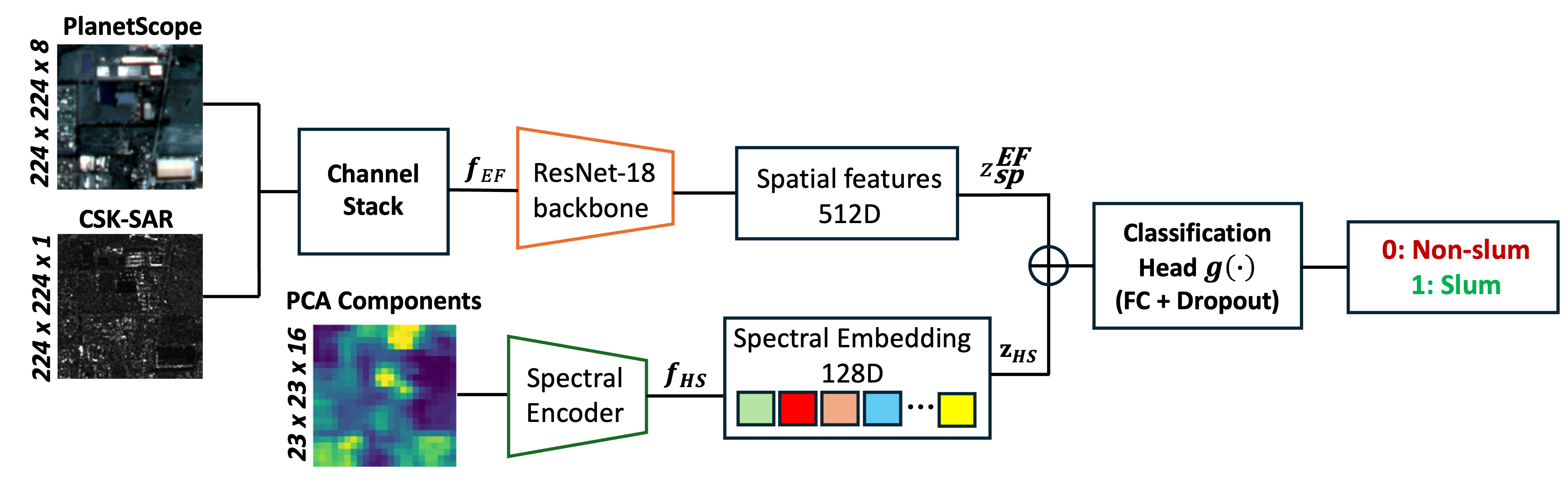} \\
        \small (a) Early fusion (EF) \\[1mm]
        \includegraphics[width=0.92\textwidth]{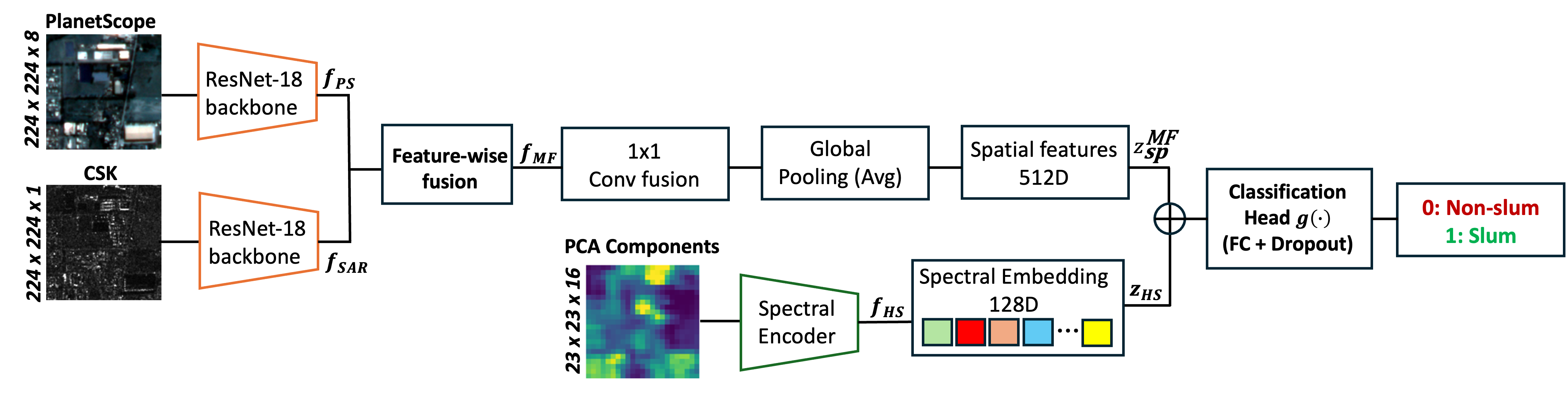} \\
        \small (b) Middle fusion (MF) \\[1mm]
        \includegraphics[width=0.68\textwidth]{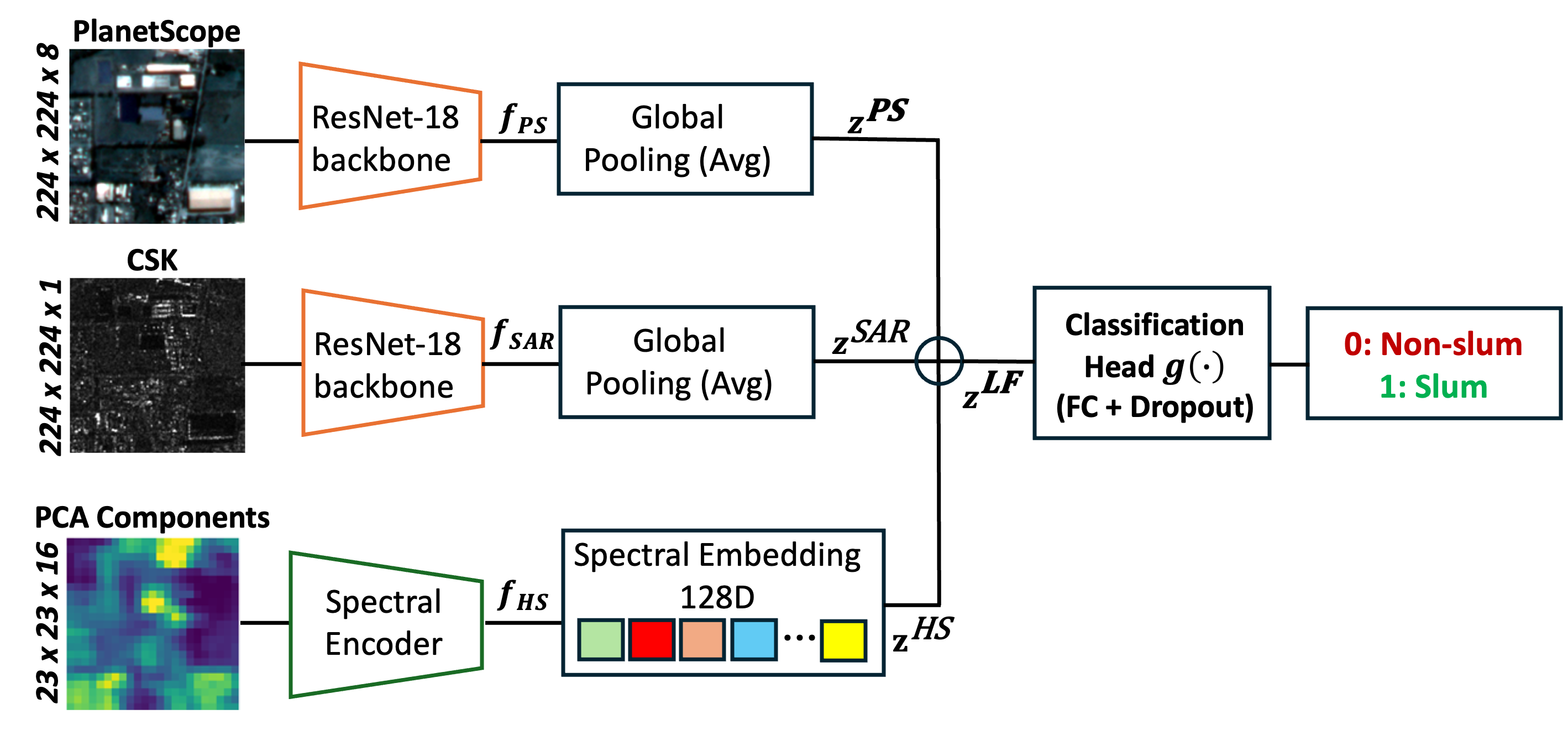} \\
        \small (c) Late fusion (LF)
    \end{tabular}
   \caption{
    Multi-sensor fusion architectures compared in this study.
    (a) EF stacks the PS and CSK inputs before a single ResNet-18 backbone.
    (b) MF processes the PS and CSK inputs with independent encoders and combines their feature maps before classification.
    (c) LF keeps the SAR and MS representations separate until feature-vector concatenation.
    In all ``+HS'' configurations, the PCA-reduced PRISMA subset is encoded as a compact spectral embedding and appended to the final feature representation before classification.
    }
    \label{fig:fusion_architectures}
\end{figure*}

\subsection{Learning Framework}
\label{subsec:learning_framework}
All experiments are based on ResNet-18 convolutional neural networks adapted to the considered input modalities and fusion configurations. For input configurations with a number of channels different from three, the original input convolution was replaced with a convolution matching the required input dimensionality: one channel for SAR, eight channels for MS, and nine channels for the EF configuration. The new convolution retained the original number of output channels, kernel size, stride, and padding. The original classification layer was replaced by an identity mapping so that the network acted as a convolutional feature extractor. Given an input sample $\mathbf{x}^{(s)}_i$ from modality $s$, the corresponding spatial feature representation is obtained as:
\begin{equation}
    \mathbf{F}^{(s)}_i =
    f_s
    \left(
    \mathbf{x}^{(s)}_i
    \right),
\end{equation}
where $f_s(\cdot)$ denotes the modality-specific ResNet-18 encoder and $\mathbf{F}^{(s)}_i \in \mathbb{R}^{C \times H \times W}$ is the extracted feature map.

A compact latent representation is then obtained through global average pooling (GAP):
\begin{equation}
    \mathbf{z}^{(s)}_i =
    \mathrm{GAP}
    \left(
    \mathbf{F}^{(s)}_i
    \right),
\end{equation}
where $\mathbf{z}^{(s)}_i \in \mathbb{R}^{C}$ denotes the pooled feature vector associated with modality $s$.

Depending on the considered configuration, the final representation $\mathbf{z}_i$ corresponds either to a single-modality feature vector or to the fusion of multiple modality-specific representations. The resulting vector is processed by a classification head composed of dropout regularisation followed by a fully connected layer:
\begin{equation}
    o_i =
    g
    \left(
    \mathbf{z}_i
    \right),
\end{equation}
where $g(\cdot)$ denotes the classification head and $o_i \in \mathbb{R}$ is the output logit associated with patch $i$.

The model outputs a scalar logit $o_i$. During training, the networks are optimised using weighted binary cross-entropy with logits, applied directly to $o_i$:
\begin{equation}
    \mathcal{L} =
    -
    \frac{1}{N}
    \sum_{i=1}^{N}
    \left[
        w^{+} y_i \log(\sigma(o_i))
        +
        (1-y_i)\log(1-\sigma(o_i))
    \right].
\end{equation}

For inference and mapping, the corresponding slum-likelihood score is obtained by applying a sigmoid activation:
\begin{equation}
    \hat{p}_i =
    \sigma(o_i).
\end{equation}
To account for class imbalance, the positive class is weighted more strongly in the loss. For each OOF iteration, the positive-class weight $w^{+}$ is computed using only the training folds as
\begin{equation}
    w^{+} =
    \frac{N^{-}_{\mathrm{train}}}{N^{+}_{\mathrm{train}}},
\end{equation}
where $N^{-}_{\mathrm{train}}$ and $N^{+}_{\mathrm{train}}$ are the numbers of ReNaBaP-negative and ReNaBaP-positive patches in the training set, respectively.

\subsection{PRISMA HS Embedding}
\label{subsec:prisma_embedding}

PRISMA HS information is incorporated as a compact spectral embedding rather than being directly stacked with the high-resolution MS and SAR inputs. It is therefore used as a material-sensitive spectral descriptor associated with each high-resolution patch, rather than as an additional spatial backbone. The PCA-reduced subsets retain their native spatial support during preprocessing. Before being provided to the HS encoder, however, each component is spatially averaged to obtain a compact patch-level spectral descriptor. For each sample, the PCA-reduced PRISMA subset $\tilde{\mathbf{x}}^{\mathrm{HS}}_i \in \mathbb{R}^{K \times h \times w}$, with $K=16$, is spatially averaged over its native-resolution support:

\begin{equation}
    \bar{x}^{\mathrm{HS}}_{i,k} =
    \frac{1}{hw}
    \sum_{u=1}^{h}
    \sum_{v=1}^{w}
    \tilde{x}^{\mathrm{HS}}_{i,k,u,v},
    \qquad k=1,\ldots,K .
\end{equation}

This operation produces a patch-level spectral descriptor $\bar{\mathbf{x}}^{\mathrm{HS}}_i \in \mathbb{R}^{16}$, which summarises the mean PCA-reduced HS response over the same geographic footprint covered by the corresponding PS and CSK patches. The spatially averaged PCA descriptor is then processed by a lightweight encoder composed of two 1D convolutional layers, followed by a fully connected projection into a 128-dimensional HS embedding:

\begin{equation}
    \mathbf{z}^{\mathrm{HS}}_i =
    f_{\mathrm{HS}}
    \left(
    \bar{\mathbf{x}}^{\mathrm{HS}}_i
    \right)
    \in \mathbb{R}^{128}.
\end{equation}

This embedding provides material-sensitive spectral information from PRISMA. When PRISMA is included, $\mathbf{z}^{\mathrm{HS}}_i$ is concatenated with the SAR, MS or fused spatial feature vector before classification.

\subsection{SAR- and MS-Based Configurations}
\label{subsec:single_sensor_baselines}

Two single-sensor baselines are first defined using either CSK SAR or PS MS patches as the only high-resolution input. In the SAR-only configuration, the model receives $\mathbf{x}^{\mathrm{SAR}}_i$, while in the MS-only configuration it receives $\mathbf{x}^{\mathrm{PS}}_i$. The corresponding feature vector is obtained as
\begin{equation}
\mathbf{z}^{(s)}_i =
\mathrm{GAP}
\left(
f_s
\left(
\mathbf{x}^{(s)}_i
\right)
\right),
\qquad
s \in \{\mathrm{SAR},\mathrm{PS}\}.
\end{equation}

Two additional single-spatial-backbone configurations incorporate the PRISMA embedding described in Section~\ref{subsec:prisma_embedding}. In these cases, CSK or PS remains the only high-resolution spatial input processed by a ResNet-18 backbone, while PRISMA provides a compact spectral embedding associated with the same geographic footprint. The final representation is therefore
\begin{equation}
\mathbf{z}_i =
\mathrm{Concat}
\left(
\mathbf{z}^{(s)}_i,
\mathbf{z}^{\mathrm{HS}}_i
\right).
\end{equation}

Otherwise, $\mathbf{z}_i=\mathbf{z}^{(s)}_i$. The SAR-only and MS-only models provide the single-sensor baselines, whereas the SAR+HS and MS+HS variants quantify the contribution of PRISMA while retaining a single high-resolution spatial backbone.

\subsection{Multi-Sensor Fusion Strategies}
\label{subsec:fusion_strategies}

Three strategies are evaluated for SAR--MS integration: EF, MF, and LF. These strategies differ in the stage at which the SAR and MS representations are combined. The EF, MF, and LF terminology therefore refers specifically to the integration of the two high-spatial-resolution modalities. In configurations denoted by ``+HS'', the PRISMA embedding described in Section~\ref{subsec:prisma_embedding} is appended to the final feature representation and is not involved in the EF/MF/LF integration stage. The three architectures are schematically shown in Fig.~\ref{fig:fusion_architectures}(a)--(c).

\subsubsection{Early Fusion}
\label{subsubsec:early_fusion}

In EF, CSK and PS patches are concatenated to form a joint input representation:
\begin{equation}
    \mathbf{x}^{\mathrm{EF}}_i =
    \mathrm{Concat}_{c}
    \left(
    \mathbf{x}^{\mathrm{SAR}}_i,
    \mathbf{x}^{\mathrm{PS}}_i
    \right),
\end{equation}
where $\mathrm{Concat}_{c}(\cdot)$ denotes channel-wise concatenation. The resulting tensor therefore has nine input channels and is processed by a single ResNet-18 encoder:
\begin{equation}
    \mathbf{z}^{\mathrm{EF}}_{\mathrm{sp},i} =
    \mathrm{GAP}
    \left(
    f_{\mathrm{EF}}
    \left(
    \mathbf{x}^{\mathrm{EF}}_i
    \right)
    \right).
\end{equation}

If PRISMA is included, the final EF representation is
\begin{equation}
    \mathbf{z}^{\mathrm{EF}}_i =
    \mathrm{Concat}
    \left(
    \mathbf{z}^{\mathrm{EF}}_{\mathrm{sp},i},
    \mathbf{z}^{\mathrm{HS}}_i
    \right).
\end{equation}

Otherwise,
\begin{equation}
\mathbf{z}^{\mathrm{EF}}_i =
\mathbf{z}^{\mathrm{EF}}_{\mathrm{sp},i}.
\end{equation}

EF allows SAR backscatter and MS reflectance to interact from the first convolutional layers. This tests whether SAR sensitivity to surface roughness and built-up geometry can complement the spectral and textural information provided by MS imagery.

\subsubsection{Middle Fusion}
\label{subsubsec:middle_fusion}

In MF, SAR and PS inputs are first processed by independent ResNet-18 encoders:
\begin{equation}
    \mathbf{F}^{\mathrm{SAR}}_i =
    f_{\mathrm{SAR}}
    \left(
    \mathbf{x}^{\mathrm{SAR}}_i
    \right),
    \qquad
    \mathbf{F}^{\mathrm{PS}}_i =
    f_{\mathrm{PS}}
    \left(
    \mathbf{x}^{\mathrm{PS}}_i
    \right).
\end{equation}

The resulting feature maps are concatenated along the channel dimension and projected through a $1 \times 1$ convolution:
\begin{equation}
    \mathbf{F}^{\mathrm{MF}}_i =
    \phi_{1 \times 1}
    \left(
    \mathrm{Concat}_{c}
    \left(
    \mathbf{F}^{\mathrm{SAR}}_i,
    \mathbf{F}^{\mathrm{PS}}_i
    \right)
    \right),
\end{equation}
where $\phi_{1 \times 1}(\cdot)$ denotes the feature-level fusion operator. The fused spatial representation is then obtained by a GAP layer:
\begin{equation}
    \mathbf{z}^{\mathrm{MF}}_{\mathrm{sp},i} =
    \mathrm{GAP}
    \left(
    \mathbf{F}^{\mathrm{MF}}_i
    \right).
\end{equation}

When PRISMA is used, the final MF representation is
\begin{equation}
    \mathbf{z}^{\mathrm{MF}}_i =
    \mathrm{Concat}
    \left(
    \mathbf{z}^{\mathrm{MF}}_{\mathrm{sp},i},
    \mathbf{z}^{\mathrm{HS}}_i
    \right).
\end{equation}
Otherwise, $\mathbf{z}^{\mathrm{MF}}_i=\mathbf{z}^{\mathrm{MF}}_{\mathrm{sp},i}$.

MF allows SAR and MS data to learn modality-specific spatial features before interaction. The $1 \times 1$ convolution then learns a compact joint representation while keeping the fused features spatially aligned.

\subsubsection{Late Fusion}
\label{subsubsec:late_fusion}

In LF, SAR and PS inputs are encoded independently by two separate ResNet-18 encoders up to the feature-vector level:
\begin{equation}
    \mathbf{z}^{\mathrm{SAR}}_i =
    \mathrm{GAP}
    \left(
    f_{\mathrm{SAR}}
    \left(
    \mathbf{x}^{\mathrm{SAR}}_i
    \right)
    \right),
\end{equation}
\begin{equation}
    \mathbf{z}^{\mathrm{PS}}_i =
    \mathrm{GAP}
    \left(
    f_{\mathrm{PS}}
    \left(
    \mathbf{x}^{\mathrm{PS}}_i
    \right)
    \right).
\end{equation}

The SAR--MS spatial representation is obtained by concatenating the modality-specific feature vectors:
\begin{equation}
\mathbf{z}^{\mathrm{LF}}_{\mathrm{sp},i} =
\mathrm{Concat}
\left(
\mathbf{z}^{\mathrm{SAR}}_i,
\mathbf{z}^{\mathrm{PS}}_i
\right).
\end{equation}

When PRISMA is included, the final LF representation is obtained by appending the HS embedding:
\begin{equation}
\mathbf{z}^{\mathrm{LF}}_i =
\mathrm{Concat}
\left(
\mathbf{z}^{\mathrm{LF}}_{\mathrm{sp},i},
\mathbf{z}^{\mathrm{HS}}_i
\right).
\end{equation}

Otherwise,
\begin{equation}
\mathbf{z}^{\mathrm{LF}}_i =
\mathbf{z}^{\mathrm{LF}}_{\mathrm{sp},i}.
\end{equation}

LF is the most conservative strategy, because SAR and MS information are kept separate until the final representation stage. This design is suitable when the input modalities differ in physical meaning and statistical properties, as in the case of MS reflectance, SAR backscatter and HS descriptors.

\subsection{Training, Evaluation and City-Wide Mapping}
\label{subsec:training_evaluation_mapping}

All model configurations are trained and evaluated using the four-fold spatial CV scheme described in Section~\ref{subsubsec:spatial_cv}. At each iteration, three spatial folds are used for training and the remaining fold is held out for evaluation. Models are trained for 25 epochs per fold with a batch size of 32. Optimisation is performed using AdamW with a learning rate of $10^{-4}$ and a weight decay of $10^{-4}$. All ResNet-18 backbones are initialised from ImageNet-pretrained weights, and a dropout rate of 0.2 is used in the classification head.

For each model configuration, the OOF slum-likelihood scores from the four iterations are aggregated for patch-level evaluation and for reconstructing city-wide slum-likelihood maps.

\section{Results and Discussion}
\label{sec:results_discussion}
This section presents the experimental results of the proposed multi-sensor framework. The analysis is organised in five steps. First, patch-level classification performance is reported for the SAR- and MS-based configurations and the multi-sensor fusion strategies. Second, the corresponding city-wide slum-likelihood maps are analysed qualitatively. Third, detections outside the ReNaBaP inventory are examined using the CBA vulnerability layer, to assess whether they are spatially consistent with broader vulnerable neighbourhoods rather than merely representing false positives. Fourth, the continuous slum-likelihood scores are aggregated at neighbourhood scale to evaluate their ability to separate high- and low-vulnerability CBA areas. Finally, a complementary thermal analysis is used to support the physical interpretation of the ReNaBaP settlements.

\begin{table*}[t]
\centering
\caption{
Patch-level classification performance of SAR- and MS-based configurations and multi-sensor fusion strategies under spatial OOF evaluation. PRISMA HS information is included as a PCA-reduced spectral embedding when ``+HS'' is indicated. The best value for each metric is highlighted in bold.
}
\label{tab:classification_results}
\scriptsize
\setlength{\tabcolsep}{4.2pt}
\renewcommand{\arraystretch}{1.18}
\resizebox{\textwidth}{!}{%
\begin{tabular}{lcccccccccc}
\toprule
\multirow{2}{*}{Metric} 
& \multicolumn{4}{c}{SAR- and MS-based configurations} 
& \multicolumn{6}{c}{Multi-sensor fusion strategies} \\
\cmidrule(lr){2-5} \cmidrule(lr){6-11}
& SAR & SAR+HS & MS & MS+HS 
& EF & EF+HS & MF & MF+HS & LF & LF+HS \\
\midrule
Precision 
& 0.3590 & 0.2623 & 0.4103 & 0.4267 
& 0.2426 & 0.2828 & 0.3913 & 0.4737 & 0.4000 & \textbf{0.5455} \\

Recall    
& 0.4746 & 0.5424 & 0.5424 & 0.5424 
& 0.6942 & \textbf{0.6949} & 0.6102 & 0.3051 & 0.6441 & 0.6102 \\

F1-score  
& 0.4088 & 0.3536 & 0.4672 & 0.4776 
& 0.3596 & 0.4020 & 0.4768 & 0.3711 & 0.4935 & \textbf{0.5760} \\

AUROC     
& 0.8198 & 0.8757 & 0.8634 & 0.8837 
& 0.8676 & \textbf{0.8929} & 0.8880 & 0.8298 & 0.8826 & 0.8674 \\

Accuracy  
& 0.9472 & 0.9237 & 0.9524 & 0.9544 
& 0.9048 & 0.9205 & 0.9485 & 0.9602 & 0.9492 & \textbf{0.9654} \\

Kappa     
& 0.3817 & 0.3182 & 0.4427 & 0.4541 
& 0.3209 & 0.3674 & 0.4511 & 0.3516 & 0.4683 & \textbf{0.5581} \\
\bottomrule
\end{tabular}%
}
\end{table*}

\begin{figure*}[t]
    \centering
    \includegraphics[width=\textwidth]{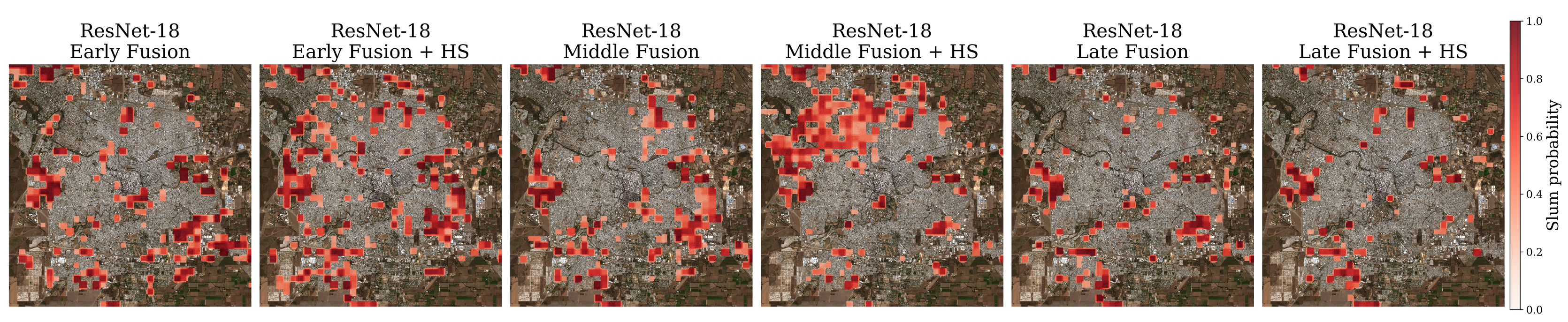}
    \caption{
    City-wide slum-likelihood maps obtained with the different ResNet-18 fusion strategies, with and without PRISMA HS support. Warmer colours indicate higher slum-likelihood scores. The comparison highlights the different spatial behaviours of the EF, MF, and LF models.
    }
    \label{fig:citywide_all_models}
\end{figure*}

\subsection{Patch-Level Classification Performance}
\label{subsec:classification_results}

Table~\ref{tab:classification_results} reports the patch-level performance obtained under the four-fold spatial OOF evaluation protocol. The comparison includes the CSK SAR-only and PS MS-only baselines, their corresponding SAR+HS and MS+HS single-spatial-backbone configurations, and the EF, MF, and LF fusion configurations, with the latter evaluated both with and without PRISMA HS support.

For each configuration, the model produces an OOF slum-likelihood score for every retained patch. AUROC is computed directly from the continuous scores and therefore does not require a binary decision threshold. For the remaining metrics, the OOF predictions from the four held-out folds are combined into a single set, and the threshold that maximises the F1-score on this pooled OOF set is selected for each model configuration. Precision, recall, F1-score, accuracy and Cohen's kappa therefore represent performance at an operating point optimised on the pooled OOF predictions.

The SAR-only and MS-only baselines show that PS MS imagery provides a stronger standalone representation than CSK SAR. The MS baseline reaches an F1-score of 0.4672 and a Cohen's kappa of 0.4427, compared with 0.4088 and 0.3817 for SAR-only input. This behaviour is expected, since high-resolution MS imagery directly captures spectral reflectance patterns, vegetation scarcity, exposed soil and fine-scale urban texture, all of which are visually associated with informal-settlement morphology. SAR alone remains informative, but its lower precision (0.3590) suggests that backscatter responses associated with roughness, geometry and built-up structure are less specific when used without MS context.

The addition of PRISMA HS embeddings has a non-uniform effect. For the MS baseline, the inclusion of HS information slightly improves performance, increasing the F1-score from 0.4672 to 0.4776 and the AUROC from 0.8634 to 0.8837. However, HS information does not consistently improve all configurations. This behaviour suggests that HS descriptors mainly provide complementary spectral information related to surface materials, whose effectiveness depends on how they are integrated with the high-resolution spatial representations.

Among the fusion strategies, EF obtains the highest recall, reaching 0.6949 in the EF+HS configuration, but also the lowest precision values (0.2426 without HS and 0.2828 with HS). This behaviour suggests that EF makes the model more sensitive to slum-like patterns, but less selective, resulting in more false positives. MF provides a relatively balanced trade-off between precision and recall when HS information is not used, achieving an F1-score of 0.4768 and an AUROC of 0.8880, but becomes substantially more conservative after the inclusion of HS embeddings, with recall decreasing to 0.3051.

LF achieves the best overall trade-off between precision and recall. In particular, LF+HS obtains the highest precision (0.5455), F1-score (0.5760), accuracy (0.9654) and Cohen's kappa (0.5581) among all tested configurations. Although its recall (0.6102) is lower than EF, the substantially higher precision indicates a better balance between sensitivity and false-positive control. These results suggest that preserving modality-specific representations before final integration is beneficial when combining physically different observations such as MS reflectance, SAR backscatter and HS descriptors.

\begin{figure*}[t]
    \centering
    \includegraphics[width=\textwidth]{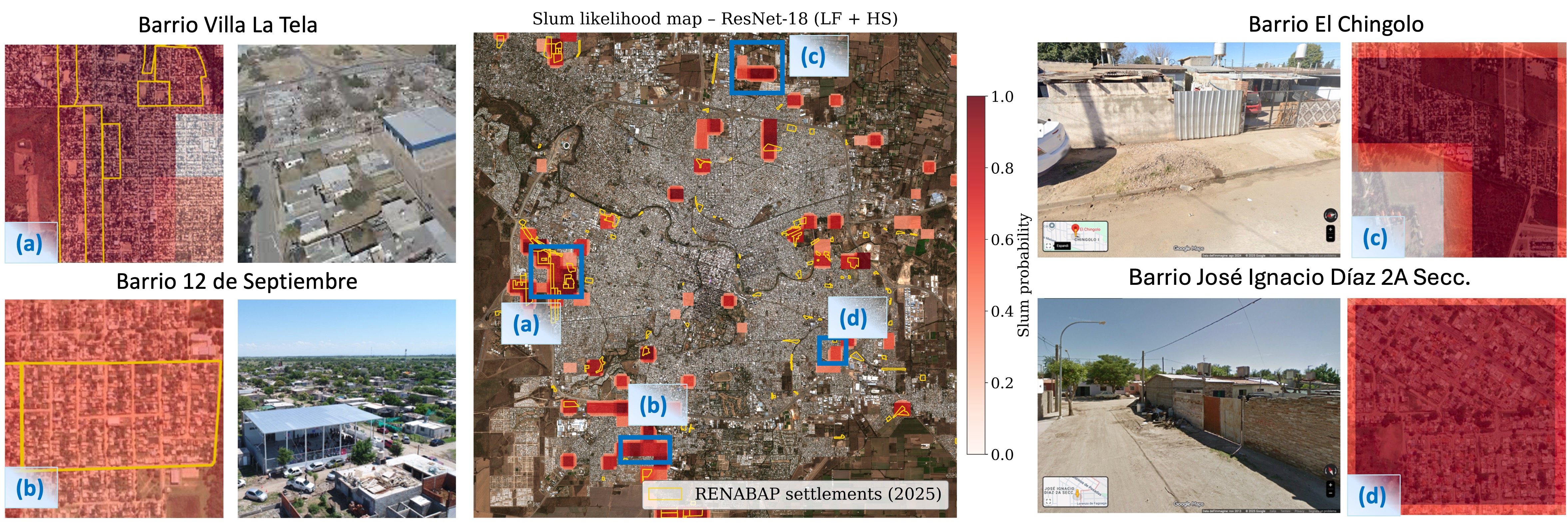}
    \caption{
        City-wide slum-likelihood map obtained with the ResNet-18 LF+HS configuration. ReNaBaP settlements are shown as yellow polygons. The insets show examples of detected official settlements, including (a) Barrio Villa La Tela and (b) Barrio 12 de Septiembre, and areas with high slum-likelihood scores outside the adopted ReNaBaP inventory, including (c) Barrio El Chingolo and (d) Barrio Jos\'e Ignacio D\'iaz 2A Secc.
    }
    \label{fig:citywide_lfhs_examples}
\end{figure*}

\subsection{City-Wide Slum-Likelihood Mapping}
\label{subsec:citywide_mapping}

The OOF patch-level predictions are projected back onto the PS reference grid to generate city-wide slum-likelihood maps. For each model configuration, the prediction assigned to patch $i$, denoted as $\hat{p}_i$, is associated with the geographic footprint of the corresponding PS patch. Since neighbouring patches partially overlap, the final slum-likelihood score assigned to each map pixel is obtained through a weighted aggregation of all patch predictions covering that pixel.

Let $\mathcal{P}(q)$ be the set of patches whose footprint includes pixel $q$, and let $w_i(q)$ be the spatial weight assigned to pixel $q$ within patch $i$. The final slum-likelihood map is computed as:
\begin{equation}
    P(q) =
    \frac{
    \sum_{i \in \mathcal{P}(q)}
    w_i(q)\hat{p}_i
    }{
    \sum_{i \in \mathcal{P}(q)}
    w_i(q)
    }.
\end{equation}
In this study, $w_i(q)$ is defined through a smooth radial kernel centred on each patch, giving a higher weight to the central part of the patch and a lower weight to the borders. This reduces block-like artefacts and produces smoother score transitions across overlapping patch footprints.

Thus, the city-wide map is reconstructed from the OOF predictions used for patch-level evaluation.

Fig.~\ref{fig:citywide_all_models} compares the city-wide slum-likelihood maps obtained with the different ResNet-18 fusion configurations. These maps provide a spatial interpretation of the quantitative results reported in Table~\ref{tab:classification_results}. EF and EF+HS combinations produce widespread activations across the city, including several scattered high-score patches in peripheral and non-contiguous areas. This behaviour is consistent with their high recall but limited precision, suggesting a tendency to over-detect slum-like patterns.

MF produces less spatially scattered hotspots than EF, suggesting that learning separate SAR and MS features before fusion reduces some of the scattered activations observed with input-level fusion. However, when HS embeddings are added, the slum-likelihood map becomes more diffuse, with broader activations over parts of the central and northern urban fabric. This suggests that the HS descriptors introduce additional spectral context, but in this configuration they do not improve the spatial selectivity of the detections.

By contrast, LF and especially the LF+HS configuration produce more selective slum-likelihood maps. High-score regions are less spatially scattered and are concentrated around ReNaBaP settlements or areas visually consistent with vulnerable settlement patterns. This behaviour is coherent with the stronger precision, F1-score and kappa obtained by LF+HS, and supports its selection as the reference configuration for the following interpretation analyses.

Fig.~\ref{fig:citywide_lfhs_examples} shows the LF+HS slum-likelihood map with the official ReNaBaP polygons overlaid, together with representative local examples. Cases (a) and (b), corresponding to Barrio Villa La Tela and Barrio 12 de Septiembre, show correctly detected settlements, with high-score hotspots overlapping the reference polygons. These areas exhibit dense low-rise urban fabric, irregular local morphology and heterogeneous roofing patterns that are visually consistent with vulnerable settlement forms.

Cases (c) and (d), corresponding to Barrio El Chingolo and Barrio José Ignacio Díaz 2A Secc., receive high slum-likelihood scores despite being outside the adopted ReNaBaP reference inventory. If evaluated only against ReNaBaP, these detections would therefore be interpreted as false positives. However, the map and the Google Street View insets suggest that these areas may correspond to broader socio-spatial vulnerability patterns not fully represented by the official inventory.

These observations suggest that the model output should not be interpreted exclusively as a binary indication of registered \emph{barrios populares}. Rather, the slum-likelihood map may also highlight urban configurations associated with vulnerability beyond the boundaries of the adopted inventory. Accordingly, some detections counted as false positives with respect to ReNaBaP may occur in vulnerable or transitional areas not represented in official settlement registers.

\subsection{Contextual Assessment of ReNaBaP-Based False Positives Using the CBA Vulnerability Layer}
\label{subsec:contextual_analysis}

As discussed above, the primary quantitative evaluation considers ReNaBaP as the reference inventory. Consequently, predicted positive areas located outside the adopted ReNaBaP polygons are formally treated as false positives with respect to this reference. However, this designation indicates disagreement with the adopted inventory and does not necessarily imply the absence of broader urban vulnerability.

Some detections with high slum-likelihood scores outside ReNaBaP are located in urban areas exhibiting spatial and morphological characteristics consistent with vulnerable settlement patterns, as illustrated in Fig.~\ref{fig:citywide_lfhs_examples}(c)--(d). To investigate the broader urban context of these detections, the LF+HS slum-likelihood map was compared with the Municipality of Córdoba vulnerability layer (CBA).

Unlike ReNaBaP, which delineates officially registered \emph{barrios populares}, the CBA dataset represents broader neighbourhood-scale vulnerability conditions and distinguishes between high- and low-vulnerability areas. The CBA layer was therefore used exclusively as an external contextual dataset for the interpretation of the model outputs. 

\begin{figure}[t]
\centering
\includegraphics[width=\columnwidth]{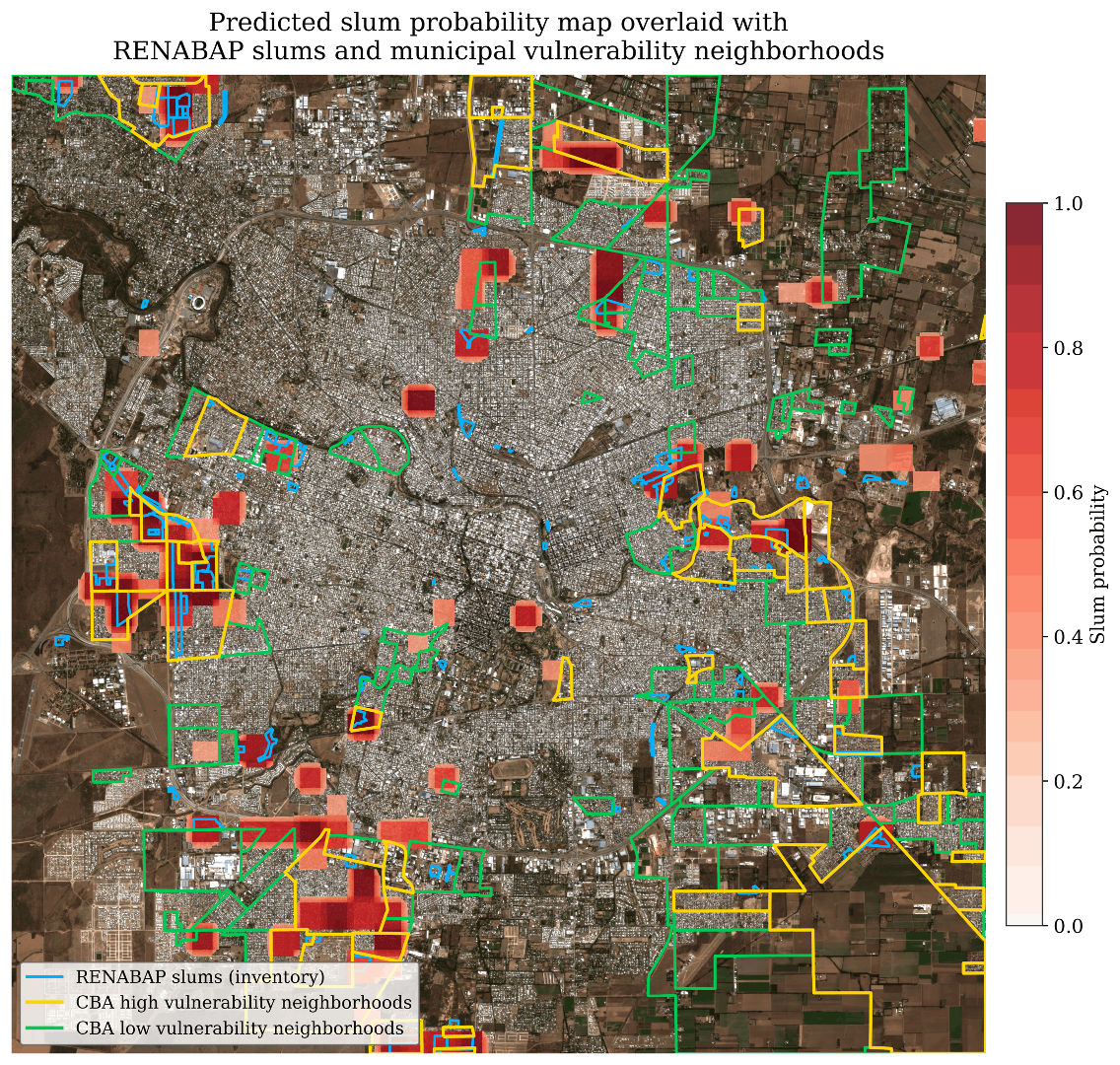}
\caption{
LF+HS slum-likelihood map overlaid with the adopted ReNaBaP inventory and the Municipality of Córdoba vulnerability layer. ReNaBaP polygons are shown in blue, high-vulnerability CBA neighbourhoods in yellow and low-vulnerability neighbourhoods in green. Several areas with high slum-likelihood scores outside ReNaBaP fall within broader areas identified by the municipality as vulnerable. The CBA polygons are used only for contextual interpretation and do not constitute additional slum reference labels.
}
\label{fig:lfhs_cba_overlay}
\end{figure}

Fig.~\ref{fig:lfhs_cba_overlay} shows the LF+HS slum-likelihood map together with the adopted ReNaBaP polygons and the vulnerable neighbourhoods identified in the CBA layer.

For this spatial-overlap analysis, the LF+HS slum-likelihood map was binarised using the global decision threshold adopted for the corresponding OOF evaluation. Let $\widehat{Y}_{\tau}$ denote the resulting binary prediction map and let $R$ denote the rasterised ReNaBaP inventory on the same spatial grid. The mask of predicted positive locations outside ReNaBaP was defined as

\begin{equation}
M_{\mathrm{FP}}
=
\left\{
q :
\widehat{Y}_{\tau}(q)=1
\ \land
R(q)=0
\right\},
\end{equation}

where $q$ denotes a location on the PS reference grid. These locations are false positives only with respect to the adopted ReNaBaP inventory.

The high- and low-vulnerability CBA polygons were rasterised onto the same grid, producing the masks $C_{\mathrm{H}}$ and $C_{\mathrm{L}}$, respectively. The overall CBA vulnerability mask was defined as

\begin{equation}
C_{\mathrm{CBA}}
=
C_{\mathrm{H}} \cup C_{\mathrm{L}}.
\end{equation}

The proportion of the ReNaBaP-based false-positive area falling within CBA vulnerable neighbourhoods was then computed as

\begin{equation}
\rho_{\mathrm{CBA}}
=
\frac{
\left|
M_{\mathrm{FP}} \cap C_{\mathrm{CBA}}
\right|
}{
\left|M_{\mathrm{FP}}\right|
}.
\end{equation}

Because all masks are represented on the same grid, $\rho_{\mathrm{CBA}}$ can equivalently be interpreted as the proportion of mapped false-positive pixels or corresponding mapped area overlapping the CBA layer. Nevertheless, the resulting pixel values are derived from spatially aggregated patch-level predictions and should not be interpreted as independent pixel-wise classifications.

The analysis shows that approximately $60\%$ of the area formally identified as false positive with respect to ReNaBaP falls within neighbourhoods represented in the CBA layer. This substantial overlap provides contextual support for the qualitative examples presented in Fig.~\ref{fig:citywide_lfhs_examples}(c)--(d) and indicates that part of the apparent false-positive response is spatially associated with broader municipal vulnerability patterns not represented in the adopted ReNaBaP inventory.

These detections remain false positives under the ReNaBaP-based evaluation and are not reassigned as true positives. Since the CBA layer represents broad neighbourhood-level vulnerability rather than informal-settlement boundaries, it is used only for contextual interpretation and not to recompute the classification metrics or confirm previously unmapped ReNaBaP settlements. Nevertheless, the overlap indicates that a substantial portion of the mapped false-positive area outside ReNaBaP occurs within areas identified by the municipality as vulnerable.

\subsection{Vulnerability-Discrimination Analysis}
\label{subsec:vulnerability_discrimination}

Building on the previous contextual analysis, we further examined whether the LF+HS slum-likelihood map is consistent with the high- and low-vulnerability classes defined by the CBA layer.

For each CBA neighbourhood, the maximum slum-likelihood score was extracted from the LF+HS OOF map and used as an indicator of the highest local model response within that area.

\begin{figure}[t]
    \centering
    \includegraphics[width=\columnwidth]{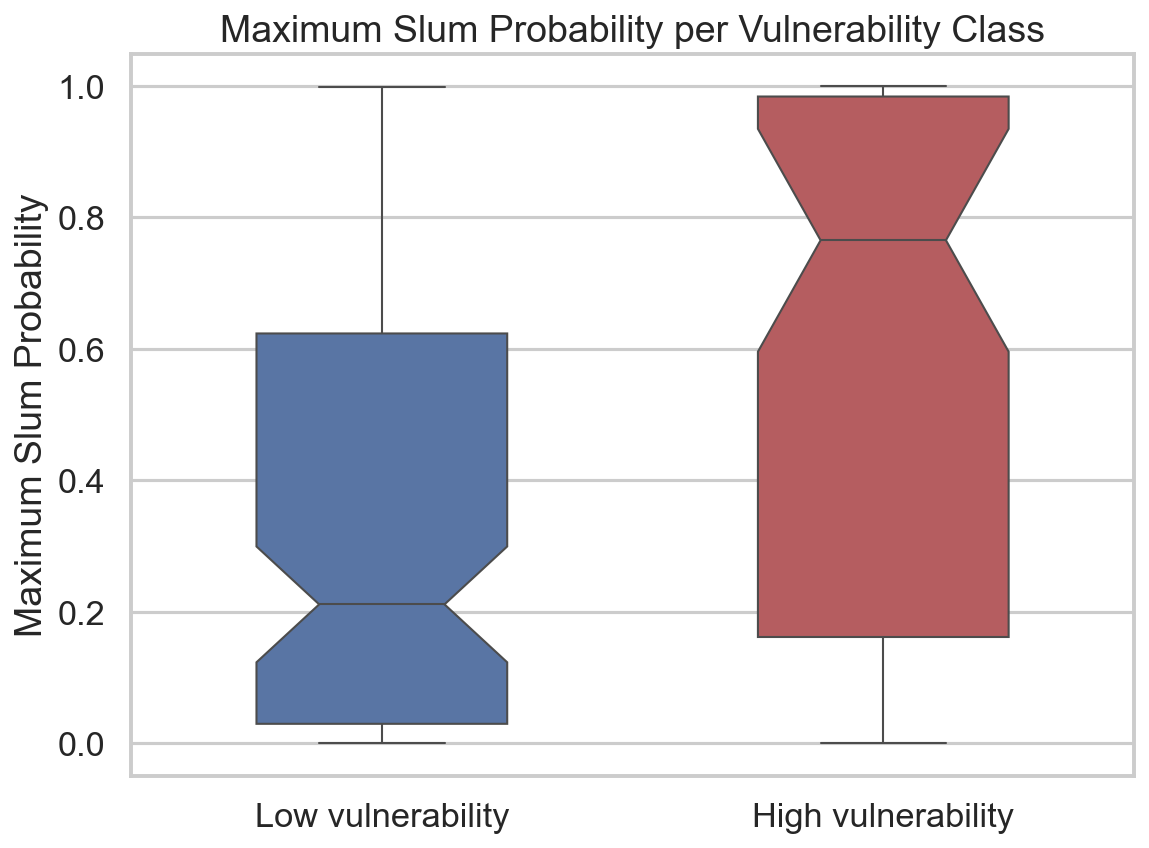}
    \caption{
    Distribution of maximum LF+HS slum-likelihood values within CBA high- and low-vulnerability neighbourhoods.
    }
    \label{fig:vulnerability_boxplot}
\end{figure}

Fig.~\ref{fig:vulnerability_boxplot} compares the distribution of maximum slum-likelihood values between high- and low-vulnerability CBA neighbourhoods. The distributions show a shift towards higher values in high-vulnerability neighbourhoods. High-vulnerability areas tend to exhibit substantially higher slum-likelihood scores, with a higher median and a stronger concentration toward the upper part of the slum-likelihood-score range. By contrast, low-vulnerability neighbourhoods are generally associated with lower maximum scores, although a limited number of high-score outliers are still present.

This behaviour suggests that the LF+HS slum-likelihood map is not only consistent with the spatial distribution of ReNaBaP settlements, but is also sensitive to broader neighbourhood-scale vulnerability conditions captured by the municipal layer.

\begin{figure}[t]
    \centering
    \includegraphics[width=\columnwidth]{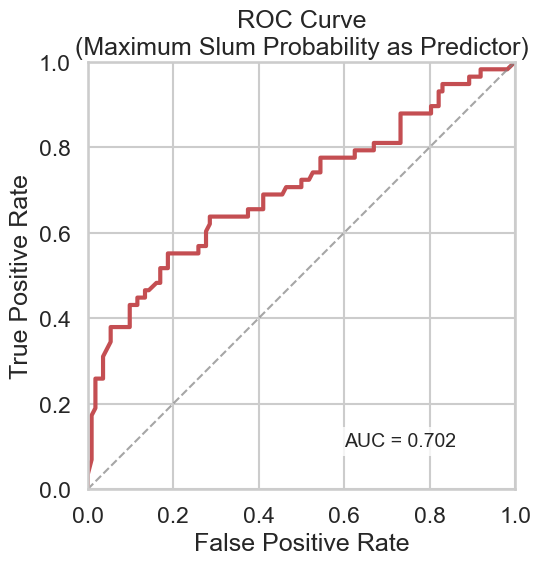}
    \caption{
    ROC curve obtained using the maximum LF+HS slum-likelihood value within each CBA neighbourhood to discriminate between high- and low-vulnerability classes.
    }
    \label{fig:vulnerability_roc}
\end{figure}

A Receiver Operating Characteristic (ROC) analysis was performed using the maximum slum-likelihood score extracted from the LF+HS OOF map within each CBA neighbourhood as predictor and the municipal vulnerability class as binary target.  The resulting ROC curve is shown in Fig.~\ref{fig:vulnerability_roc}.

The curve lies consistently above the random baseline and yields an Area Under the Curve (AUC) of 0.702, indicating a moderate discrimination in the analysed CBA dataset. In practical terms, this means that high-vulnerability neighbourhoods tend to receive higher slum-likelihood values than low-vulnerability ones in a substantial fraction of pairwise comparisons.

Although the separation is not perfect, these results provide additional evidence that the proposed framework captures spatial patterns associated with broader urban vulnerability conditions. Overall, the LF+HS slum-likelihood map shows spatial patterns associated not only with ReNaBaP settlements but also with broader patterns of urban vulnerability.

\subsection{Thermal Interpretation During a Heatwave Event}
\label{subsec:thermal_analysis}

As a final interpretation step, we analysed whether ReNaBaP settlements also show distinct thermal behaviour during extreme heat conditions. For this purpose, a Landsat~8 OLI/TIRS Collection~2 Level-2 surface-temperature product acquired over Córdoba on 03 January 2022 was used to estimate surface temperature during a summer heatwave event.

The comparison was performed locally around each ReNaBaP settlement. For every polygon, a 250~m surrounding buffer was generated, excluding the settlement itself. In this way, surface temperature inside each settlement could be compared with that of its immediate surroundings, reducing the influence of larger-scale temperature variations.

\begin{figure}[t]
    \centering
    \includegraphics[width=\columnwidth]{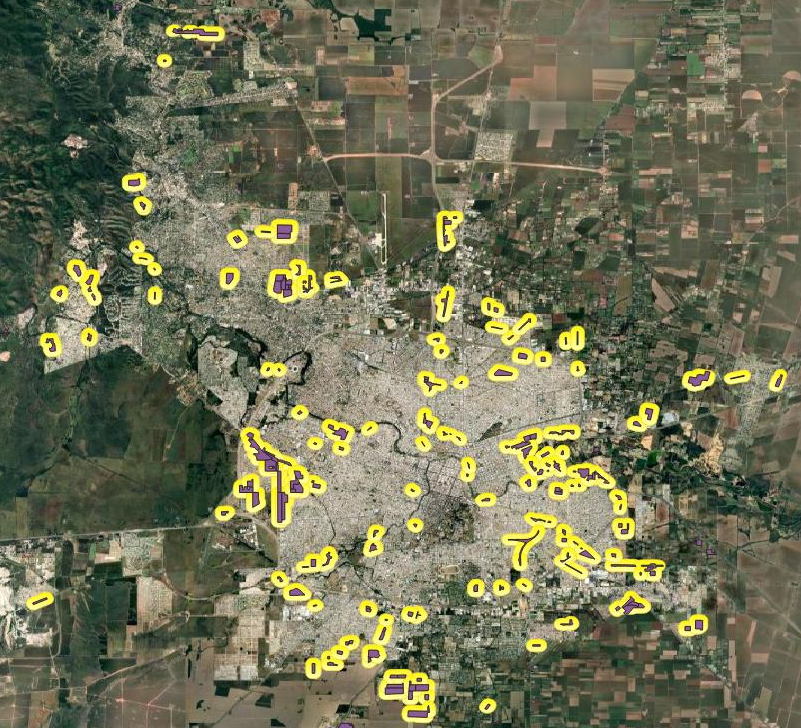}
    \caption{
    Thermal analysis setup for ReNaBaP settlements and their surrounding areas. Official ReNaBaP polygons are shown in purple, while yellow areas indicate the 250~m buffer zones used for local temperature comparison after excluding the ReNaBaP polygons.
    }
    \label{fig:thermal_analysis}
\end{figure}

Fig.~\ref{fig:thermal_analysis} shows the ReNaBaP polygons and the corresponding buffer zones used for this analysis.

The results show that ReNaBaP settlements are warmer than their local surroundings. The mean surface temperature is $42.60^\circ$C inside the settlements and $41.96^\circ$C in the surrounding buffer areas, with an average difference of about $+0.63^\circ$C. This difference is statistically significant according to a Welch two-sample t-test ($t = 22$, $p < 2.2 \times 10^{-16}$).

Although the temperature increase is moderate, it is meaningful because the comparison is local. It suggests that ReNaBaP settlements may be associated with localised surface-heat amplification during extreme temperature events. This is consistent with their dense built-up structure, limited vegetation, heterogeneous roofing materials, and reduced local cooling capacity, and agrees with previous studies reporting elevated heat exposure and thermal stress in informal settlements \cite{Scott2017NairobiHeat,Xia2026}.


\subsection{Limitations and Discussion}
\label{subsec:limitations}

The experimental results highlight several relevant aspects concerning the contribution of the different sensing modalities, the interpretation of the resulting slum-likelihood maps, and the main limitations of the proposed framework.

First, the comparison among fusion strategies suggests that preserving modality-specific representations before integration is beneficial when combining observations acquired by heterogeneous EO sensors. The LF+HS configuration achieved the best overall trade-off among precision, F1-score, accuracy, and Cohen’s kappa. This behaviour is likely related to the substantially different physical characteristics of the considered inputs. PS imagery primarily captures MS reflectance and fine-scale spatial texture, CSK SAR backscatter is sensitive to surface geometry, roughness, and structural scattering mechanisms, while the PRISMA-derived descriptors encode lower-resolution spectral variability associated with surface composition.

In the EF configuration, SAR and MS data are combined at the input level and are therefore forced to interact from the first convolutional layers. This may cause modality-specific statistical properties to be mixed before sufficiently modality-specific representations are learned, potentially explaining the comparatively high recall but limited precision observed for EF. By contrast, LF preserves independent feature-extraction pathways for the SAR and MS modalities and integrates their representations only after modality-specific encoding. This architecture appears to provide a more selective representation of settlement-related spatial patterns.

The experiments also indicate that HS descriptors provide complementary spectral information when combined with high-resolution MS and SAR features, rather than acting as the primary source of discrimination. Their contribution appears to enrich the learned representation with additional material-related variability, thereby increasing the complementarity of the final multimodal representation. This interpretation is physically plausible because the native 30~m spatial resolution of PRISMA is too coarse to resolve narrow streets, individual roofs, or other fine-scale morphological characteristics of informal settlements. Instead, PRISMA mainly contributes spatially aggregated spectral information potentially related to surface composition, vegetation cover, exposed soil, and material heterogeneity. The performance of LF+HS therefore suggests that material-sensitive HS descriptors can provide useful contextual support when integrated with higher-resolution spatial representations.

An important limitation of the present study concerns the temporal mismatch among the EO datasets. The PS and CSK acquisitions refer to 2025, whereas the PRISMA scene was acquired in 2021 and the Landsat-8 thermal image in 2022. The proposed framework should therefore not be interpreted as a temporally aligned multi-sensor acquisition setup.

This limitation is partly mitigated by the specific role assigned to each dataset. PRISMA is not used to reconstruct fine-scale settlement morphology, but to provide compact material-sensitive spectral descriptors over broader urban areas. Similarly, Landsat-8 thermal information is not included in DL training, threshold selection, or city-wide inference. Urban environments may evolve between acquisition dates, and changes in buildings, land cover, vegetation, and surface materials may affect the correspondence among the datasets. Nevertheless, the present analysis focuses primarily on consolidated settlements and relatively persistent urban characteristics, such as compact built-up fabric, material heterogeneity, limited vegetation, and exposed surfaces, rather than on short-term changes affecting individual structures. The objective is therefore not to model temporal dynamics, but to investigate whether heterogeneous EO observations provide complementary information related to broader settlement and vulnerability patterns.

The thermal analysis must be interpreted separately from the LF+HS mapping results. The temperature comparison was performed exclusively within the official ReNaBaP polygons and their surrounding buffer areas. It was not conducted over all slum-likelihood areas identified by LF+HS and, in particular, did not evaluate model detections located outside the adopted ReNaBaP inventory. Consequently, the observed surface-temperature difference cannot be interpreted as an independent thermal validation of the LF+HS detections or as evidence that all model-derived hotspots experience enhanced thermal exposure.

Rather, the thermal analysis provides complementary information concerning the ReNaBaP settlements. During the analysed heatwave event, the surfaces within the ReNaBaP polygons exhibited, on average, higher temperatures than their immediate surroundings. This result suggests that the ReNaBaP settlements may be associated with localised surface-heat amplification, potentially related to dense built-up conditions, roofing and paving materials, limited vegetation, and reduced evaporative cooling. However, the present analysis does not establish which specific physical or socio-environmental factors caused the observed temperature difference.

Several additional limitations affect the interpretation of the thermal results. First, the relatively coarse spatial resolution of Landsat-8 TIRS data does not allow temperature variability to be resolved at the scale of individual buildings or narrow urban structures. Second, the analysis is based on a single acquisition during a specific heatwave event and therefore does not characterise seasonal, interannual, or diurnal thermal dynamics. Third, the temporal mismatch between the thermal image and the high-resolution EO observations prevents a direct correspondence between the measured surface-temperature patterns and the urban conditions represented in the PS and CSK imagery. The thermal comparison should therefore be regarded as an exploratory analysis of the ReNaBaP settlements rather than as a comprehensive assessment of thermal vulnerability or a validation of the LF+HS predictions.

Another important consideration concerns the interpretation of the ReNaBaP inventory itself. ReNaBaP provides an official institutional reference with georeferenced settlement boundaries and therefore constitutes a suitable baseline for supervised learning and quantitative evaluation. However, it should not be interpreted as a complete or exhaustive representation of all forms of urban vulnerability in Córdoba. Official inventories may be influenced by administrative recognition procedures, update frequency, data-collection practices, and the institutional criteria adopted for settlement inclusion.

Consequently, model detections located outside the adopted ReNaBaP polygons are formally classified as false positives in the primary evaluation, although some may occur within urban areas exhibiting morphological or socio-spatial characteristics associated with broader vulnerability. The comparison with the external CBA vulnerability layer provides contextual support for this interpretation. Nevertheless, the CBA polygons represent broader neighbourhood-level vulnerability conditions rather than detailed boundaries of informal settlements. Their overlap with LF+HS detections therefore cannot be regarded as confirmation of previously unmapped settlements or used to replace the ReNaBaP-based evaluation. The CBA comparison should instead be interpreted as a complementary contextual analysis showing that part of the apparent false-positive area occurs within urban areas identified as vulnerable by the municipality.

The use of patch-level labels introduces an additional source of uncertainty. Each patch is assigned a binary label according to its overlap with the ReNaBaP polygons, whereas the urban scene contained within a patch may include a heterogeneous mixture of settlement types, buildings, vegetation, roads, vacant land, and other surface classes. Although ambiguous patches were excluded during dataset construction, retained positive and negative samples may still contain substantial internal heterogeneity. Moreover, during city-wide inference, patch-level predictions are spatially aggregated to generate a continuous slum-likelihood map. The resulting pixel-level values should therefore be interpreted as spatially smoothed slum-likelihood estimates derived from overlapping patch predictions, rather than as direct pixel-wise classifications of individual buildings or land-cover elements.

Finally, the proposed analysis is limited to a single urban context. Informal settlements are inherently context-dependent, and their morphology, density, construction materials, vegetation patterns, and spatial organisation may vary substantially across cities and countries. Sensor acquisition geometry, climate, seasonality, urban structure, and the characteristics of the available reference inventories may also influence model behaviour. Consequently, the results obtained for Córdoba should not be considered directly transferable to other urban systems without additional adaptation and independent validation.

Future work should therefore investigate the transferability of the proposed framework across different cities and geographical contexts. This could include cross-city and cross-country validation, domain-adaptation strategies, uncertainty-aware modelling, and the use of larger and more heterogeneous reference datasets integrating institutional, municipal, and community-level information. Further developments could also explore more advanced and physically interpretable HS representations, multi-temporal MS and SAR observations, and the direct integration of thermal variables into the learning framework. Such an extension would require temporally aligned thermal observations and independent reference information to assess whether model detections outside ReNaBaP are also systematically associated with distinct environmental-exposure conditions.

\section{Conclusions}
\label{sec:conclusions}
This work presented a multi-sensor DL framework for slum-likelihood mapping in Córdoba, Argentina, integrating high-resolution multispectral imagery, SAR data and HS observations and evaluating the models using four geographically partitioned folds. The resulting maps were subsequently interpreted using the external CBA vulnerability layer and a separate thermal analysis. The experimental results showed that PS MS imagery achieved the strongest single-modality performance, while CSK SAR provided complementary structural information related to urban geometry and built-up texture.
PRISMA HS observations mainly acted as complementary spectral information alongside the higher-resolution MS and SAR representations. Among the evaluated fusion strategies, the LF+HS configuration achieved the best overall balance between classification performance and spatial selectivity, suggesting that preserving modality-specific features during feature extraction supports a more selective slum-likelihood representation when combining physically heterogeneous EO observations.

The resulting city-wide slum-likelihood maps showed spatial agreement with several ReNaBaP settlements, while also highlighting additional areas with high slum-likelihood scores outside the adopted inventory. The comparison with the external CBA vulnerability layer suggested that part of the apparent false-positive area is spatially consistent with vulnerable urban contexts not fully represented in the official settlement inventory. Furthermore, the neighbourhood-scale analysis showed that higher LF+HS slum-likelihood scores were generally associated with neighbourhoods classified as more vulnerable, extending the interpretation beyond the boundaries of officially mapped settlements.

The additional thermal analysis provided a complementary physical interpretation of the ReNaBaP settlements. These settlements exhibited significantly higher surface temperatures than their immediate surroundings during the analysed heatwave event, suggesting a potential association with localised surface-temperature increase. Since the thermal comparison was restricted to ReNaBaP polygons, this result should not be interpreted as a thermal validation of the additional areas detected by the LF+HS model.

These findings indicate that multi-sensor EO fusion can support both the mapping of ReNaBaP settlements and the interpretation of broader urban vulnerability patterns. Future work should assess the transferability of the proposed framework across different cities, integrate uncertainty-aware modelling strategies, and exploit more heterogeneous reference inventories combining institutional and community-level information. Further developments could also investigate the direct integration of thermal and other physically meaningful environmental variables within multi-modal learning frameworks for urban vulnerability analysis.

\section*{Acknowledgments}
This research was carried out within the framework of a PhD funded by NextGenerationEU, Action 4, DM n. 118, 02/03/2023. COSMO-SkyMed Products \textcopyright{} Italian Space Agency (ASI) were accessed under a license to use granted by ASI within the Project Card GLOURBEXT. PRISMA data were also provided by ASI under a specific data access agreement. PlanetScope imagery \textcopyright{} Planet Labs PBC was provided by Planet Labs PBC; access to PlanetScope data was made available through the University of Pavia institutional subscription and the Planet Education and Research Program. This research was also carried out within the HEATCORDOBA project (Future Earth--ESA-545-2025-02).

\begingroup
\sloppy
\bibliographystyle{IEEEtran}
\bibliography{references}
\endgroup

\begin{minipage}[t]{0.47\textwidth}
\begin{IEEEbiography}[{\includegraphics[width=1in,height=1.25in,clip,keepaspectratio]{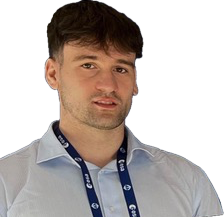}}]{Luigi Russo}
Student Member, IEEE, earned his master's degree (cum laude) in Electronic Engineering for Automation and Telecommunications from the University of Sannio, Benevento, Italy, in 2023. He is currently working toward the Ph.D. degree in Electronics, Computer Science and Electrical Engineering at the University of Pavia, Italy, in collaboration with the Italian Space Agency (ASI), Rome, Italy. He was the winner of the IEEE GRS Italy Chapter Frank Marzano Award for one of the top three Master's theses in Geosciences and Remote Sensing. He has coauthored and presented papers at several prestigious conferences and received the Best Poster Award in Urban and Data Analysis as a young scientist at the 2024 European Space Agency (ESA)-Dragon Symposium. His research interests include the application of artificial intelligence and remote sensing to the analysis of urban and environmental tasks, with particular emphasis on damage mapping and exposure assessment following natural disasters and extreme events.
\end{IEEEbiography}
\end{minipage}

\begin{minipage}[t]{0.47\textwidth}
\begin{IEEEbiography}[{\includegraphics[width=1.05in,height=1.20in,clip,keepaspectratio]{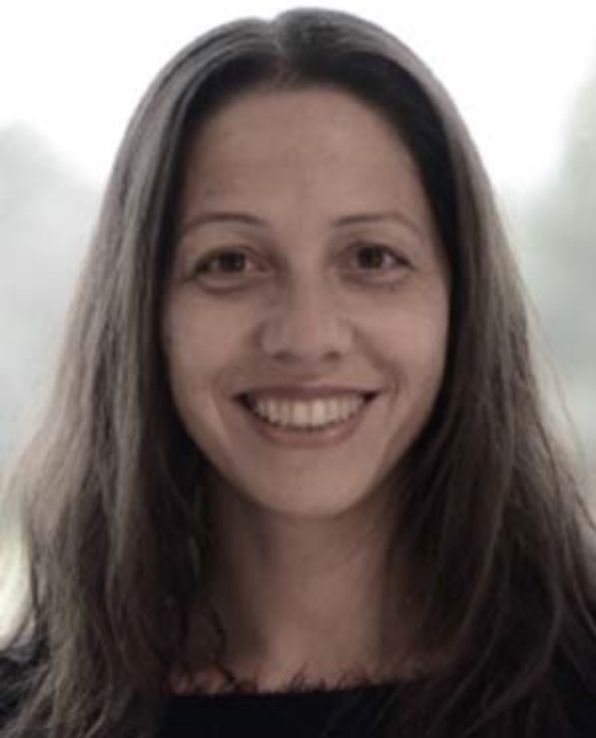}}]{Anabella Ferral}
Senior Member, IEEE, received the B.Sc. degree in chemistry and the Ph.D. degree in chemistry, in the field of physical chemistry, from the National University of Córdoba, Córdoba, Argentina, in 1998 and 2005, respectively, and the master’s degree in space applications of early warning and response to emergencies from Gulich Institute (IG), Córdoba, and the Faculty of Mathematics, Astronomy, Physics and Computing (FAMAF), National University of Córdoba in 2013.
From 2015 to 2020, she was the Head of that program. She is currently a Professor and Researcher with the IG with a position at the National Council for Scientific and Technological Research (CONICET) and has national and international peer-reviewed publications in the areas of computational chemistry and environmental monitoring by remote sensing techniques. In the last years, she has been researching and advising Ph.D. and M.Sc. students in topics related to remote sensing applied to the development and modeling of environmental quality indicators, within the framework of the SDG 2030. Dr. Ferral was the recipient of a scholarship from the National Commission of Space Activities (CONAE) to obtain the master’s degree.
\end{IEEEbiography} 
\end{minipage}


\begin{minipage}[t]{0.47\textwidth}
\begin{IEEEbiography}[{\includegraphics[width=1.05in,height=1.20in,clip,keepaspectratio]{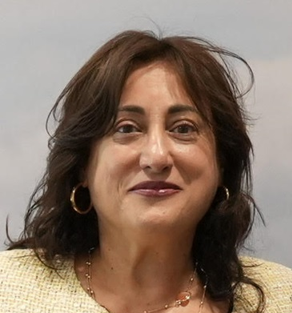}}]{Silvia Liberata Ullo}
IEEE Senior Member, President of IEEE AESS Italy Chapter, Industry Liaison for IEEE Joint ComSoc/VTS Italy Chapter since 2018. Member of the Image Analysis and Data Fusion Technical Committee (IADF TC) of the IEEE Geoscience and Remote Sensing Society (GRSS) since 2020. GRSS Europe Liaison since January 2024. Editor in Chief of IET Image Processing. Graduated with laude in 1989 in Electronic Engineering at the University of Naples (Italy), pursued the M.Sc. in Management at MIT (Massachusetts Institute of Technology, USA) in 1992. Researcher and teacher since 2004 at University of Sannio, Benevento (Italy). Member of Academic Senate and PhD Professors’ Board. Courses: Signal theory and elaboration, Telecommunication networks (Bachelor program); Earth monitoring and mission analysis Lab (Master program), Optical and radar remote sensing (Ph.D. program). Authored 140+ research papers, co-authored many book chapters and served as editor of two books. Associate Editor of relevant journals (IEEE TGRS, IEEE JSTARS, IEEE GRSL, MDPI Remote Sensing, IET Image Processing, Springer Arabian Journal of Geosciences). Research interests: signal processing, radar systems, sensor networks, smart grids, remote sensing, satellite data analysis, machine learning and quantum ML applied to remote sensing.
\end{IEEEbiography} 
\end{minipage}


\begin{minipage}[t]{0.47\textwidth}
\begin{IEEEbiography}[{\includegraphics[width=1in,height=1.25in,clip,keepaspectratio]{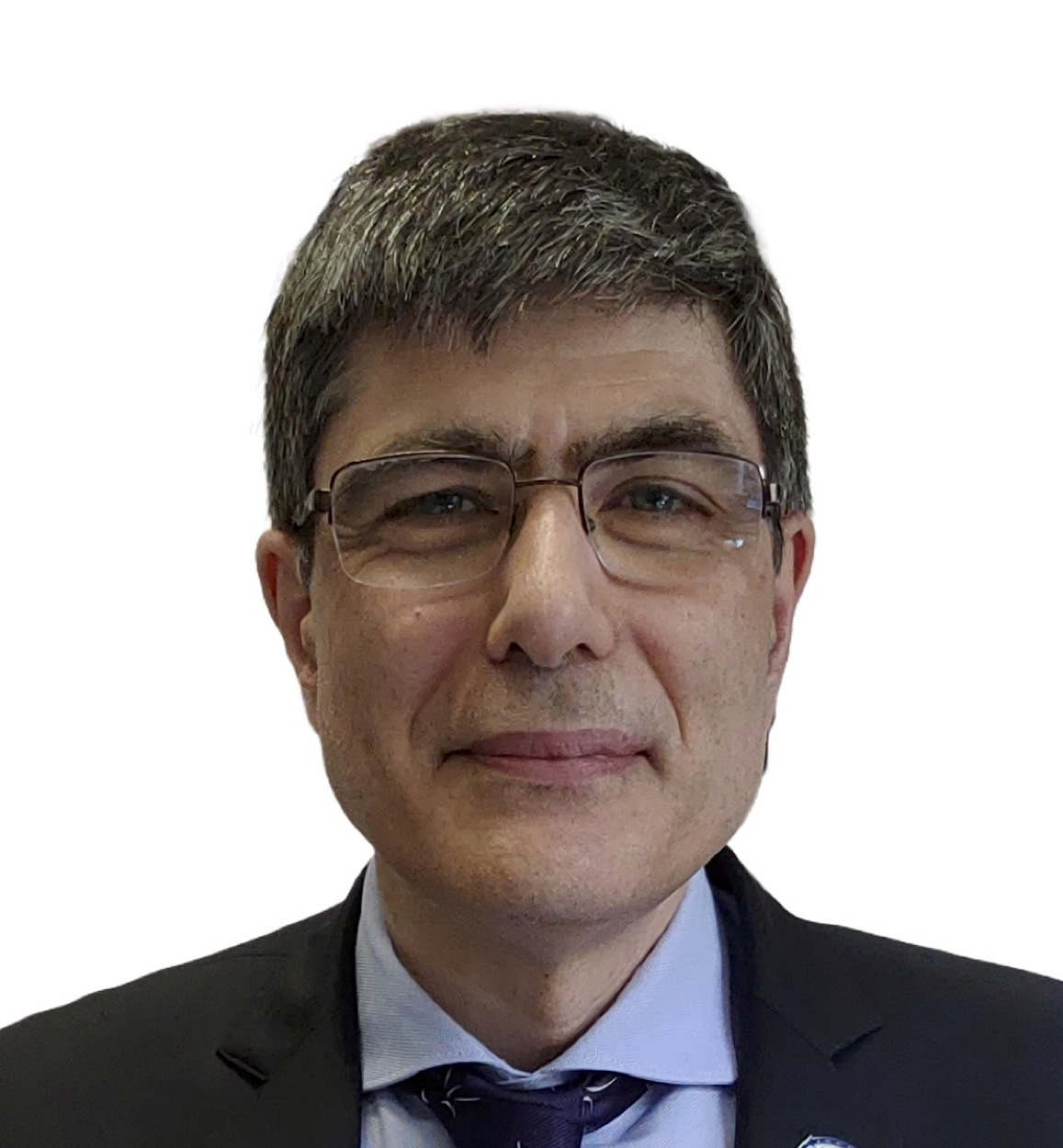}}]{Paolo Gamba} is Professor at the University of Pavia, Italy in the Telecommunications and Remote Sensing Laboratory. He received the Laurea degree in Electronic Engineering “cum laude” from the University of Pavia, Italy, in 1989, and the Ph.D. in Electronic Engineering from the same University in 1993.
He served as Editor-in-Chief of the IEEE Geoscience and Remote Sensing Letters from 2009 to 2013 and of the IEEE Geoscience and Remote Sensing Magazine in 2023-2024. He was Chair of the Data Fusion Committee of the IEEE Geoscience and Remote Sensing Society (GRSS) from October 2005 to May 2009. He has been elected in the GRSS AdCom from 2014 to 2022 and served as GRSS President from 2019 to 2020. He also served as Technical Co-Chair of the 2010, 2015 and 2020 IGARSS conferences, in Honolulu (Hawaii), Milan (Italy), and on-line, respectively.
He is Fellow of IEEE, IAPR, AAIA and the Academia Europaea. He has been invited to give keynote lectures and tutorials on several occasions about urban remote sensing, data fusion, EO data for physical exposure and risk management. He published more than 210 papers in international peer-review journals.
\end{IEEEbiography}
\end{minipage}

\end{document}